\documentclass[lettersize,journal]{IEEEtran}
\usepackage[switch]{lineno}
\usepackage{amsmath,amsfonts}
\usepackage{algorithmic}
\usepackage{algorithm}
\usepackage{array}
\usepackage{multirow}

\usepackage[caption=false,font=normalsize,labelfont=sf,textfont=sf]{subfig}
\usepackage{textcomp}
\usepackage{stfloats}
\usepackage{url}
\usepackage{verbatim}
\usepackage{graphicx}
\usepackage{cite}
\usepackage{epsfig}
\usepackage{orcidlink}

\begin{document}
\title{A Test-Time Learning Approach to Reparameterize the Geophysical Inverse Problem with a Convolutional Neural Network}

\author{Anran Xu\,\orcidlink{0009-0001-9731-377X}, Lindsey J. Heagy\orcidlink{0000-0002-1551-5926}
\thanks{Anran Xu and Lindsey J. Heagy are with the Department of Earth, Ocean and Atmospheric Sciences, University of British Columbia, Vancouver, BC V6T 1Z4 Canada (email: anranxu@eoas.ubc.ca; lheagy@eoas.ubc.ca)}
\thanks{\textcopyright 2024 IEEE.  Personal use of this material is permitted.  Permission from IEEE must be obtained for all other uses, in any current or future media, including reprinting/republishing this material for advertising or promotional purposes, creating new collective works, for resale or redistribution to servers or lists, or reuse of any copyrighted component of this work in other works.}}



\maketitle

\begin{abstract}
Regularization is critical for solving ill-posed geophysical inverse problems. Explicit regularization is often used, but there are opportunities to explore the implicit regularization effects that are inherent in a Neural Network structure.  Researchers have discovered that the Convolutional Neural Network (CNN) architecture inherently enforces a regularization that is advantageous for addressing diverse inverse problems in computer vision, including de-noising and in-painting. In this study, we examine the applicability of this implicit regularization to geophysical inversions. The CNN maps an arbitrary vector to the model space. The predicted subsurface model is then fed into a forward numerical simulation to generate corresponding predicted measurements. Subsequently, the objective function value is computed by comparing these predicted measurements with the observed measurements. The backpropagation algorithm is employed to update the trainable parameters of the CNN during the inversion. Note that the CNN in our proposed method does not require training before the inversion, rather, the CNN weights are estimated in the inversion process, hence this is a test-time learning (TTL) approach. In this study, we choose to focus on the Direct Current (DC) resistivity inverse problem, which is representative of typical Tikhonov-style geophysical inversions (e.g. gravity, electromagnetic, etc.), to test our hypothesis. The experimental results demonstrate that the implicit regularization can be useful in some DC resistivity inversions. We also provide a discussion of the potential sources of this implicit regularization introduced from the CNN architecture and discuss some practical guides for applying the proposed method to other geophysical methods.

\end{abstract}


\section{Introduction}
\label{sec:1}
\IEEEPARstart{G}{eophysical} inversions are widely used in mining, environmental, and engineering applications \cite{ref25}, \cite{ref30}. Although forward modelling and optimization strategies have been developed for several decades, it’s still challenging to recover the subsurface structures mainly because this class of problems is highly ill-posed \cite{ref36}. For this reason, researchers put a lot of effort into exploring well-functioning regularizers for the inversion. 

A conventional approach employs the Tikhonov-regularized method where an explicit regularization term, comprised of smallness and smoothness terms, is included in the objective function \cite{ref25}, \cite{ref36}, \cite{ref37}. The $l_2$ norm is standard in the Tikhonov-regularized method \cite{ref5}, but there are works on using different norms to promote sparse or compact structures \cite{ref37}, \cite{ref21}. Researchers have also developed strategies for incorporating physical properties such as geological or petrophysical data into the Tikhonov-style geophysical geophysical inversions \cite{ref37}, \cite{ref33}, \cite{ref34}. The recent emergence of deep learning has garnered significant attention from researchers, particularly regarding the integration of machine learning algorithms into the regularization \cite{ref30}, \cite{ref17}, \cite{ref29}, \cite{ref35}.  

Supervised learning methods have been employed in scenarios where researchers have prior knowledge of the subsurface geological or petrophysical models. The synthetic datasets can be built based on the prior knowledge to train ML models \cite{ref39}, \cite{ref40}. One promising approach is to search for a solution in a latent space whose distribution contains prior information learned from the training samples \cite{ref2}. For example, in Ground-penetrating radar (GPR) inversion, Laloy \textit{et al.} \cite{ref8} and Lopez-Alvis \textit{et al.} \cite{ref11} trained Variational Autoencoders to map the distributions of the spatial patterns of the subsurface materials into simple distributions and integrated this learned prior into the objective function. In seismic full-waveform inversion (FWI), Mosser \textit{et al.} \cite{ref12} trained a generative adversarial network with the knowledge of subsurface geology. These methods predict more realistic subsurface models. The utilization of generative models requires a training dataset, which implies that we need prior knowledge of what we are looking for in the subsurface (e.g. the possible geological models or subsurface materials of the survey domain). Therefore, these approaches utilize a physically motivated prior.
Although the ML models can learn prior knowledge with a synthesis dataset, these methods are not applicable when we have limited prior information. In the past five years, self-supervised and unsupervised learning approaches have been applied to inverse problems. One main approach employs so-called test-time learning (TTL) methods which do not require any training data. Rather than adjusting the weights of a neural network using training samples, the weights of a neural network are learned at the time the neural network is applied to the inversion process. TTL was first applied in the inverse problems in computer vision (CV) \cite{ref20}. One example of the inverse problem in CV is in-painting (e.g. filling in corrupted portions of an image). This is achieved by learning the self-similarity between the corrupted image and the predicted image. Researchers have found that some neural networks, such as Convolutional Neural Networks (CNNs) and Graph Convolutional Networks (GCNs), inherently enforce a regularization that is advantageous for addressing diverse CV inverse problems or mesh restoration problems \cite{ref6}, \cite{ref20}. TTL methods have been applied to many biomedical imaging problems\cite{ref16}, \cite{ref31}. 

In geophysics, TTL was first adapted to seismic FWI inversion by integrating the ML architecture into the inverse process\cite{ref7}, \cite{ref14}, \cite{ref15}, \cite{ref18}. He \textit{et al.} \cite{ref7} and Wu \textit{et al.} \cite{ref14} reparameterized the velocity model into the CNN domain and attributed the regularization effect to the prior information learning by fitting the initial velocity models in the pre-training stage. In contrast, Zhu \textit{et al.} \cite{ref18} stated that this regularization effect comes from the CNN architectures as what Ulyanov \textit{et al.} \cite{ref20} proposed. Zhu \textit{et al.} \cite{ref18} examined this statement in the synthetic cases and found that the implicit regularization effect is robust to noisy data. Instead of having a fixed latent vector as the input in the previous works, researchers have tried replacing the latent vector with the observed seismic data and achieved better inversion results than the conventional FWI with total variation regularization \cite{ref41}. TTL has also been adapted to joint inversion of geophysical data \cite{ref42}, where researchers used a Bayesian neural network and a CNN to reparametrize the global model parameters and other geological model parameters, respectively. Utilizing trainable weights within the NN to parameterize the subsurface model worked well to regularize the seismic inverse problems, even though the prior they imposed on the solution is not a physically motivated prior. 

The physics that governs seismic FWI is the wave equation, and the strategies taken to solve these inverse problems often differ from those used in EM and potential field inversions, which are diffusive in nature. Problems such as EM and potential field inversions are typically solved using a Tikhonov-regularized method where a first-order smoothness term is always included \cite{ref47}. 
In this study, we would like to examine the differences between the conventional inversion that uses a smoothness term and an approach that leverages implicit regularization effects from a CNN. 
We choose the direct current (DC) resistivity inversion, which is one type of EM inverse problem, but the proposed method can be applied to other Tikhonov-style geophysical inversions. To the best of our knowledge, this is the first study that directly investigates the prior captured by the CNN for a DC resistivity inversion by integrating the CNN into the DC resistivity inversion pipeline in a truly unsupervised manner. There are works on applying CNNs in EM inversions \cite{ref43}, \cite{ref44}, \cite{ref45} in a purely data-driven way. Alternatively, B. Liu \textit{et al.} \cite{ref9} proposed the method PhResNet, which reconstructs the subsurface model by a U-net; they trained the U-net with a dataset that has no labels in it.
B. Liu \textit{et al.} \cite{ref9} stated that using a CNN to parameterize the inverse problem can be beneficial. However, there are two priors in PhResNet: the explicit prior learned from the training dataset and the implicit prior inherently from the CNN structure. It is unclear how the contributions from each of these two priors impact the results since both are intertwined when CNNs are trained with a dataset. 
Additionally, the generalizability of CNN models in the previous works depends on the size of the training dataset. In contrast, our proposed method does not rely on any training data, so the foundation for our work and previous works are fundamentally different.

We refer to our proposed method as the Deep Image Prior Inversion (DIP-Inv), as it was inspired by the work of \cite{ref20}, which introduced the concept of the Deep Image Prior in computer vision. We will begin with an overview of the methodology and describe the modifications we make to the pipeline of inversion. Then we will show the results of the DIP-Inv. The results illustrate that the CNN reparameterization provides useful regularization effects; therefore, DIP-Inv gives better inversion results than the conventional methods in some cases. After that, we explore different components of the CNN architecture for DIP-Inv and identify the bi-linear upsampling operation, which smooths the final predicted model efficiently in some scenarios, as one of the main sources of this implicit regularization effect. 

The main contributions of our work include: 
\begin{enumerate}
\item{Comparing inversion results obtained from conventional Tikhonov-regularized methods with sparse norms and those obtained from the DIP-Inv method on multiple synthetic examples. Our results illustrate that the implicit regularization effect inherently from CNN is beneficial in the Tikhonov-style geophysical inversions.}
\item{Examining some key choices in the CNN architecture to illustrate that the bi-linear upsampling operation is one of the main sources of this implicit regularization effect. And the dropout technique is also useful in some cases.}
\item{Discussing some practical details for connecting the forward simulation, written in SimPEG \cite{ref3}, with PyTorch \cite{ref24} to parameterize the model with a CNN.}
\end{enumerate}

\section{Methodology}
\label{sec:2}
In this section, we will first compare the conventional method and the DIP-Inv method (DIP-Inv) for solving the geophysical inverse problem. Then, we will illustrate the structure of DIP-Inv. Finally, we will explain the objective function of the DIP-Inv method in detail. Note that we only conduct 2D DC resistivity inversions in this study, but this methodology can be adapted to 3D DC resistivity inversions or other Tikhonov-style geophysical inversions. 

\subsection{Overview of the Conventional Inversion Method}
In a DC resistivity survey, transmitters inject a steady-state electrical current into the ground and the receivers measure the resulting distribution of potentials (voltages) on the surface. The observation $d^{obs} \in R^N$ consists of the potential differences between the two electrodes in each receiver and $N$ is the number of transmitter-receiver pairs. 
The Poisson equation governs the physics of the DC resistivity experiment
\begin{equation}
\label{Poisson equation}
\boldsymbol{\nabla} \cdot \sigma\boldsymbol{\nabla}\phi = -\boldsymbol{\nabla}\cdot\mathbf{j}_{source}
\end{equation}
where $\sigma$ is the electrical conductivity (which is the reciprocal of electrical resistivity $\rho=1/\sigma$), $\phi$ is the electric potential, and $\mathbf{j}_{source}$ is the source current density. We solve this equation using a finite volume method where we discretize the subsurface using a 2D tensor mesh (c.f. \cite{ref19}). The mesh consists of a core region, with uniform cells, and padding regions on the lateral boundaries and the bottom of the mesh with expanding cells to ensure boundary conditions are satisfied.  
We use SimPEG, an open-source package, to implement the forward modelling \cite{ref3}. 

In a conventional Tikhonov-style inversion, we pose an optimization problem that seeks to minimize an objective function comprised of a data misfit term, $\phi_d$, and a regularization term, $\phi_m$. For the DC resistivity problem, $m$ is chosen to be the log of the subsurface conductivity, and the inverse problem takes the form
\begin{equation}
\label{Conventional_objective}
\begin{aligned}
&\min_m \:\phi_d(m) + \beta\phi_m(m)= \\
&\min_m \:\frac{1}{2}||W_d(F(m) - d^{obs})||^2 + \beta (\alpha_s ||W_s(m-m_{ref})||^{p_s}+\\&\alpha_x \left||W_x D_x m\right||^{p_x}+\alpha_z \left||W_z D_z m\right||^{p_z}).
\end{aligned}
\end{equation}

The regularization term $\phi_m$ generally includes a smallness term, which penalizes the difference from a reference model $m_{ref}$, and a first-order smoothness term that penalizes spatial gradients in the model. The relative influence of each of these terms is weighted by the scalar parameters $\alpha_{s,x,z}$. The standard choice of norms in the regularization is $p_{s,x,z}=2$, however, different norms where $p_{s,x,z}\leq 2$ can be chosen, and we will use the implementation described in \cite{ref21} for these inversions. Optionally, sensitivity weighting can be included in the weight matrices $W_{s,x,z}$, and these have been shown to help improve the convergence of inversions where $p_{s,x,z}\leq 2$, \cite{ref21}.

The conventional approach for solving this inverse problem is by first initializing $m \in M$ as the reference model $m_{ref}$, which is generally taken to be a uniform half-space, and iteratively updating $m$ to reduce the value of the objective function. Various modifications can be made to the optimization problem in \ref{Conventional_objective} when prior geologic information is available. When minimal information is available, the first-order smoothness terms generally play an important role in obtaining geologically reasonable results (e.g. \cite{ref25}). In the work that follows, we choose inversion results with both smallness and first-order smoothness terms in the regularization as benchmarks for the conventional method. 

The inversion aims to recover a model that acceptably fits the data. With the assumption that noise is Gaussian and appropriate uncertainties have been used to define $W_d$ \cite{ref47}, an acceptable fit is often taken to be $\phi_d \approx \phi_d^* $ where $\phi_d^* = N/2$. For all of the conventional inversions that we consider in this paper, a $\beta$-cooling strategy will be adopted, which is a common practice \cite{ref47}. We first estimate an initial $\beta$ by comparing the magnitude of $\phi_d$ and $\phi_m$ when evaluated on a random model and set $\beta_0 = \beta_s \phi_d/\phi_m$ to ensure that the regularization has the most influence early in the inversion. 

We then employ an optimization method, commonly a second-order approach such as inexact Gauss Newton, to compute an update to the model \cite{ref5}. We then reduce $\beta$, reducing the influence of the regularization and allowing the data misfit term to have more influence in the solution and again compute an update to the model. In an inversion where $l_2$ norms are used in the regularization, this process is stopped when we have reached the target misfit. If using a different choice of norms in the smallness or smoothness terms of the regularization (e.g. $p_s, p_x, p_z \leq 2$), then we follow the methodology described in \cite{ref21}. First, we solve the inverse problem using $p_s, p_x, p_z = 2$ and then use this model as a starting model for the optimization problem with $p_s, p_x, p_z \leq 2$. Since the regularization is no longer convex when norms are less than 2, this is a more computationally challenging problem. To address this, the $\ell_p$ norm is approximated using a Lawson Norm \cite{ref60} that is evaluated and updated using the Iterative Reweighted Least-Squares (IRLS) algorithm \cite{ref21}. Additionally, sensitivity weighting can optionally be included \cite{ref51}, \cite{ref52}. 

\begin{figure}[h!]
\centering
\includegraphics[width=3.5in]{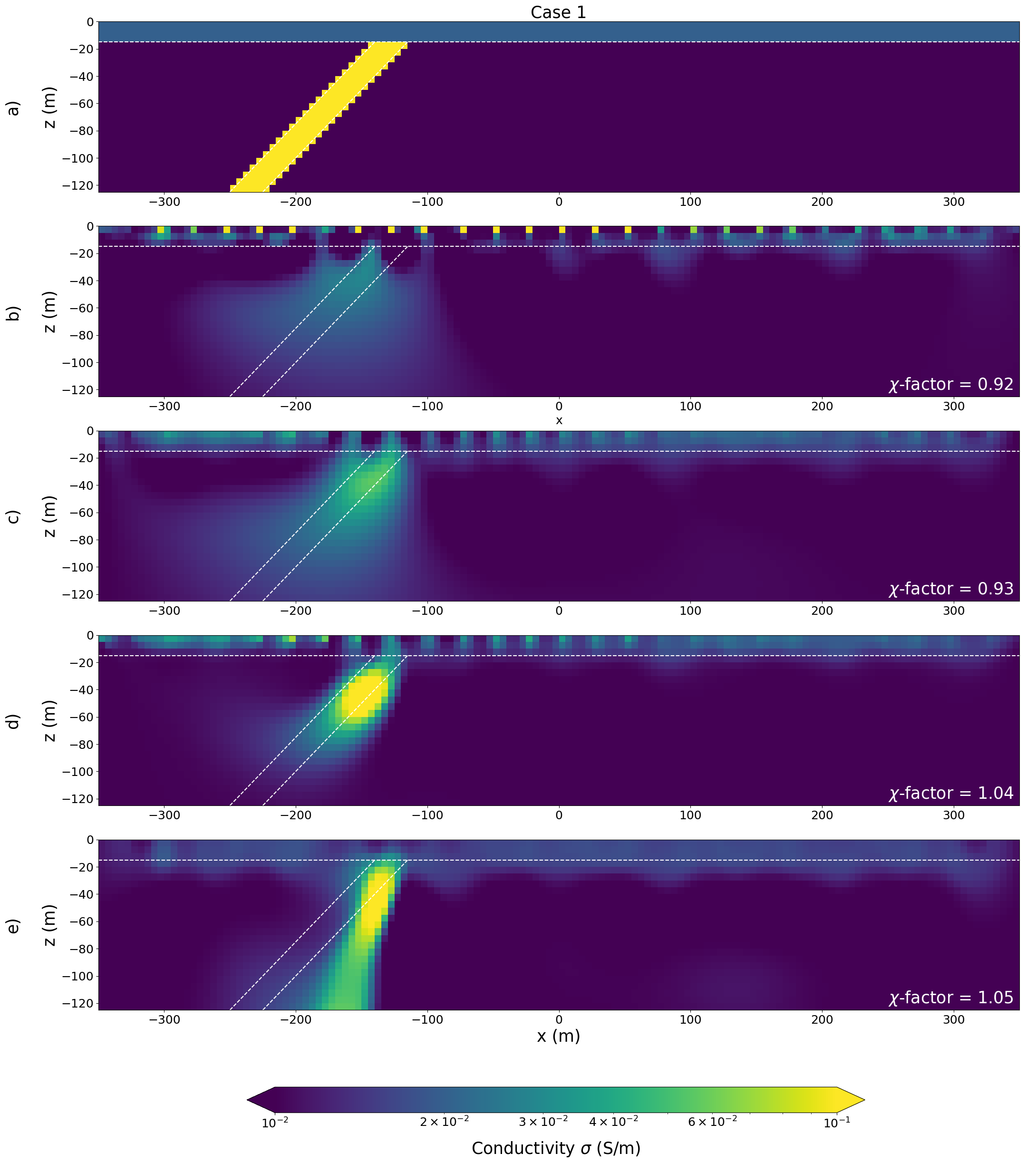}
\caption{(a) is the true conductivity model. (b) is the result of a $l_2$ inversion using only smallness. (c) is the result of a $l_2$ inversion using smallness and smoothness. (d) is the result of a $l_1$ inversion using smallness and smoothness. (e) is the result of a $l_1$ inversion using smallness, smoothness, and sensitivity weighting.}
\label{fig2.5}
\end{figure}

To illustrate the conventional approach, we consider an example of a 2-layer Earth with a dipping conductor as shown in Fig. \ref{fig2.5}(a). We generate synthetic data with 5$\%$ Gaussian noise and then conduct multiple conventional inversions. Fig. \ref{fig2.5}(b) is the result of an $l_2$ ($p_s = 2$) inversion with only the smallness term. Fig. \ref{fig2.5}(c) shows the result of an $l_2$ inversion with smallness and smoothness terms ($p_s, p_x, p_z = 2$). Fig. \ref{fig2.5}(d) shows the result when $l_1$ norms are used on the smallness and smoothness terms ($p_s, p_x, p_z = 1$). Fig. \ref{fig2.5}(e) again uses $l_1$ norms on the smallness and smoothness terms ($p_s, p_x, p_z = 1$) and additionally includes sensitivity weighting in these terms. Note that the values of $\chi$-factor ($\phi_d / \phi_d^*$, c.f. \cite{ref25}) of these inversion results are similar, but the recovered subsurface maps differ significantly, illustrating the non-uniqueness of the inverse problem. The $l_2$ ($p_s = 2$) inversion result with only a smallness term has blurred boundaries of the dike and has anomalies concentrated near the locations of the electrodes [Fig. \ref{fig2.5}(b)]. Adding a smoothness regularization reduces these near-electrodes artifacts but doesn't remove all the artifacts [Fig. \ref{fig2.5}(c)]. These $l_2$ inversion results have blurred the boundaries of the dike. In contrast, predicted models from inversions with sparse norms have sharper boundaries [Fig. \ref{fig2.5}(d)-(e)]. Fig. \ref{fig2.5}(d) is the result of a $l_1$ inversion with smallness and smoothness terms. The near-electrodes artifacts still exist. These artifacts are unwanted and problematic not only because they are geologically unreasonable but also because they will deteriorate the recovery of the anomaly structures in a deeper depth such as the dike in this case. To solve this problem, we add sensitivity weighting along with the smoothness term in a $l_1$ ($p_s, p_x, p_z = 1$) inversion, which efficiently avoids the appearance of these artifacts [\ref{fig2.5}(e)]. However, in this result, the dip angle of the recovered dike is incorrect. These inversion results show the importance of the smoothness term in obtaining a reasonable solution, the impacts of choice of the norm, and the implications of sensitivity weighting. By applying all these techniques, the predicted model has better recovery on the top layer and the shape of the dike, but there is still room for improvement as none of these solutions on their own generates fully satisfactory results. 
 
\subsection{Overview of the Proposed Parameterized Inversion Method: DIP-Inv}

\begin{figure*}[t]
\centering
\includegraphics[width=5in]{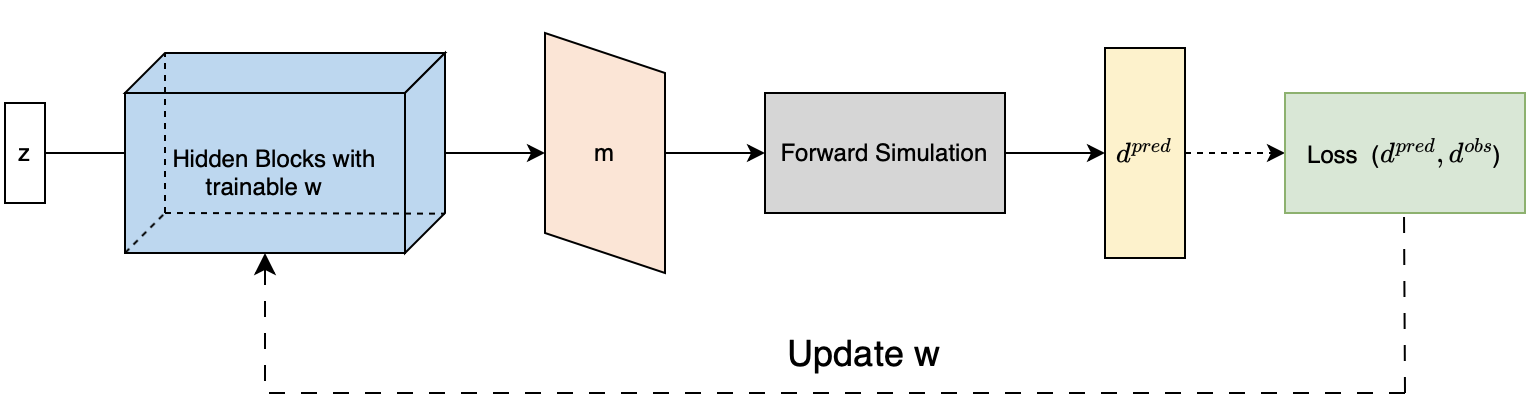}
\caption{Proposed inversion pipeline. DIP-Inv reparameterizes the mesh space into the CNN-weights space.}
\label{fig2_1}
\end{figure*}

We are interested in exploring if the implicit regularization can be advantageous in solving the Tikhonov-style geophysical inversions, so the main modification that we make to the inversion pipeline posed in \ref{Conventional_objective} is that we parameterize the model in terms of a CNN-based function. 

In our DIP-Inv method, a CNN-based function $L$, whose weights are denoted as $w$, maps an arbitrary vector $z \in R^k$ to a model $m$ in the mesh space $M \in R^{W\times H}$ [Fig. \ref{fig2_1}].

\begin{equation}
\label{Define L}
L_w(z) = m.
\end{equation}

Note that the padding cells are included in $m$. The predicted subsurface model $m$ is then fed into the forward numerical simulation implemented by SimPEG to generate the corresponding predicted measurements $d^{pred}$. The modified optimization problem that we aim to solve is 
\begin{equation}
\label{Proposed_objective}
\begin{aligned}
&\min_w (1-\beta) \phi_d + \beta \phi_m =\\&\min_w \frac{(1-\beta)}{2}||W_d(F(L_w(z)) - d^{obs})||^2 + \\&\beta ||L_w(z)-m_{ref}||^1. 
\end{aligned}
\end{equation}
 
Compared with the supervised end-to-end machine learning methods that aim to remove the physical forward simulations in the standard EM inversion, DIP-Inv imposes physical constraints in the forward simulation. Compared with the conventional Tikhonov-regularized methods, the objective function in the DIP-Inv method doesn't have the smoothness term since the smoothness constraint is implicitly included in the upsampling operator in $L$ as will be discussed in section \ref{sec:4.2}. Also, we don't use the sensitivity weighting. The backpropagation algorithm, which is implemented by PyTorch, is employed to update the trainable parameters $w$ in the function $L$. 

Since the weights $w$ are randomly initialized, the starting model $m = L_w(z)$ won't be $m_{ref}$ without pre-training. In the first stage of DIP-Inv, we update the weights $w$ with the objective function $||m-m_{ref}||^1$ to make the starting model for the second stage similar to the starting model of the conventional method.
In the second stage, we load the pre-trained weights and then update the weights with the objective function \ref{Proposed_objective}. 
Note that the implicit regularization effect does not come from the pre-training since the reference model is a uniform half-space. The first stage is mainly for saving time: the pre-trained weights in the first stage can be stored and reused for future experiments with the same reference model.

\subsection{Architecture of DIP-Inv}
\begin{figure*}[b]
\centering
\includegraphics[width=5in]{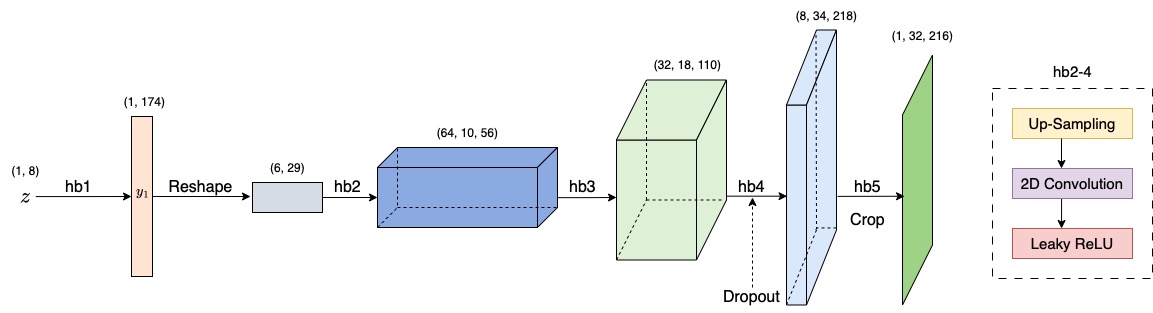}
\caption{Architecture of DIP-Inv. The structure of hidden blocks 2-4 is shown on the right (hb denotes hidden block). More details are shown in Table. \ref{table:1}.}
\label{fig2_2}
\end{figure*}

The structure of the function $L$ is illustrated in Fig. \ref{fig2_2}. The input $z$ is a fixed 1D vector sampled from the Gaussian distribution with mean 0 and standard deviation 10 and is not trainable. The first hidden block is a multi-layer perceptron (MLP) whose output is $y_1 \in R^{h}$, and the value of $h$ should be adjusted to adapt different mesh designs (i.e. dimension of $M$). Subsequently, $y_1$ is reshaped and then fed into 3 hidden blocks where each block has an upsampling layer, a 2D convolutional layer, and a Leaky ReLU activation layer. 
The last hidden layer only has a 2D convolutional layer followed by a sigmoid activation function. The output of the CNN is cropped to have the same dimension as $M$. The Sigmoid activation will squeeze the output into (0, 1). 

\begin{table}[h!]
\centering
\caption{Details of the CNN architecture implemented in vanilla DIP-Inv}
\begin{tabular}{|c|l|}
\hline
\multirow{2}{1em}{hb1} & Fully-connected layer (8, 174) \\
& LeakyReLU (negative slope 0.2)\\
\hline
 & reshape (1,1,6,29) \\
\hline
\multirow{3}{1em}{hb2} & Bi-linear upsampling ($k = 2$) \\
&Convolutional ((1,64), kernel ($3 \times 3$), stride (1), padding (0))\\
& LeakyReLU (negative slope 0.2)\\
\hline
\multirow{3}{1em}{hb3} & Bi-linear upsampling ($k = 2$) \\
&Convolutional ((64, 32), kernel ($3 \times 3$), stride (1), padding (0))\\
&LeakyReLU (negative slope 0.2)\\
\hline
\multirow{3}{1em}{hb4} & Bi-linear upsampling ($k = 2$) \\
&Convolutional ((32, 8), kernel ($3 \times 3$), stride (1), padding (0))\\
&LeakyReLU (negative slope 0.2)\\
\hline
\multirow{2}{1em}{hb5}
&Convolutional ((8, 1), kernel ($3 \times 3$), stride (1), padding (0))\\
&Sigmoid\\
\hline
output & cropped, $\times$ scale factor (problem-specific, -8 in this study)\\
\hline
\end{tabular}
\label{table:1}
\end{table}

The output of the last hidden layer is on a log-conductivity scale. All elements in the output are multiplied by -8 before being fed into the forward simulation, so the subsurface space can have conductivity values with range (exp(-8), exp(0)) S/m $\approx$ (3e-4, 1) S/m, which covers a large range of natural subsurface materials. This multiplier can be changed to different values to adapt to different conductivity ranges. 

The dropout layer can also be included in the second stage and we will examine the impact of the regularization effect from the dropout layer in section \ref{sec:4.4}. Although more advanced models have components such as inception blocks, skip connections or residual blocks, they are often based on a CNN architecture. As a result, this basic structure of CNN employed by DIP-Inv provides a good test for exploring the utility of the implicit regularization effect. 

\subsection{Necessary Explicit Regularization in DIP-Inv}
One of the main purposes of this study is to test the implicit regularization effect introduced by the CNN structures. Although one might be inclined to try to eliminate all explicit regularization, this is a very ill-posed problem (see Fig. \ref{fig2.5} for example), and we find that including the smallness term is important for this problem. Thus, we choose to analyze the impacts of the implicit regularization effects of a CNN as compared to the role of the explicit smoothness terms in the conventional approach.

The other modification that we make is in how we treat the trade-off parameter $\beta$. The cooling schedule employed in the conventional method is not suitable for the DIP-Inv method due to the difference in the convergence rate of the two methods. Compared to the conventional method, the DIP-Inv method is searching for a solution in a higher dimension space, so the number of iterations needed to reach $\chi$-factor $\approx$ 1 is larger. As a result, we need a smoother and slower cooling schedule in the DIP-Inv method.
The trade-off parameter $\beta$ in the DIP-Inv method is an exponential decay function. 
\begin{equation}
\label{dacay_curve}
\beta = e^{-\frac{t}{\tau}},
\end{equation}
where t is the index for the epochs and $\tau$ is a constant decay rate. Namely, $\beta = e^{-\frac{i}{\tau}}$ in the i-th iteration. 

\subsection{A Practical Implementation Techniques}
We choose Adam \cite{ref22} as the optimizer for the DIP-Inv. Adam updates $w$ based on the gradient of the objective function, so we need to make sure that the gradient is correctly calculated. 
Let's denote the objective function in \ref{Proposed_objective} as $\mathcal{L} = (1-\beta) \phi_d + \beta \phi_m$. Note that we use the convention of column vector in this subsection. We make updates to $w$ by considering the gradient of $\mathcal{L}$ with respect to $w$. Here, we show the derivation for the i-th element in $w$.
\begin{equation}
    \begin{split}
        \frac{\partial \mathcal{L}}{\partial w_i} &= (1-\beta) \frac{\partial \phi_d}{\partial w_i} + \beta \frac{\partial \phi_m}{\partial w_i}
        \\
        &= (1-\beta) (\nabla_m \phi_d)^T \frac{\partial m}{\partial w_i} + \beta \frac{\partial \phi_m}{\partial w_i}
        \\
        &= (1-\beta)J_v^T \frac{\partial m}{\partial w_i} + \beta \frac{\partial \phi_m}{\partial w_i}
        .\\
    \end{split}
\end{equation}
DIP-Inv uses SimPEG for the forward simulation and PyTorch for building the CNN architecture. We need to connect SimPEG, which computes $J_v = \nabla_m \phi_d$, with PyTorch which performs automatic differentiation with respect to $w$. To connect these two codes, we play a trick: We define $\mathcal{L'} =(1-\beta^t)(J_v^T m^t) + \beta^t ||m^t - m_{ref}||^1$ and $J_v^T$ is detached when calling $\mathcal{L'}.backward()$ and $optimizer.step()$ to update the weights $w$ in CNN. Although $\mathcal{L}$ and $\mathcal{L'}$ are not identical, their gradients with respect to the weights $w$ are the same. Here, we elaborate on how we implement it:
\begin{algorithm}[H]
\caption{Integrating SimPEG to the Optimization Process.}\label{alg:alg1}
\begin{algorithmic}
\STATE 
\STATE ${\textbf{Input:}}\; \tau, W, z, d^{obs}, m_{ref}, w^t, \textrm{and hyperparameters for} \;L$
\STATE $\textbf{for}\; \textrm{t = 1}\; \textbf{to}\; \textrm{...}\; \textbf{do}$
\STATE $\quad\textrm{Compute}\; \beta^t = e^{-\frac{t}{\tau}}$
\STATE $\quad\textrm{Compute}\; m^t = L_{w^t} (z)$
\STATE $\quad\textrm{Compute}\; J_v = \nabla_m \phi_d |_{m=m^t}$
\STATE $\quad\textrm{using $simulation.Jtvec(m^t, W_d^T W_d(d^{pred} - d^{obs}))$}$
\STATE $\quad\textrm{from SimPEG}$
\STATE $\quad\textrm{Compute}\;\mathcal{L'} =(1-\beta^t)(J_v^T m^t) + \beta^t ||m^t - m_{ref}||^1$
\STATE $\quad\textrm{Update}\; w_t \; \textrm{using $\mathcal{L'}.backward()$ and $optimizer.step()$} \;$
\end{algorithmic}
\label{alg1}
\end{algorithm}

\section{Trials on the Synthetic Cases}
\label{sec:3}
In this section, we will consider two synthetic cases and conduct inversions using both the conventional Tikhonov-regularized method and the proposed DIP-Inv method. The main target for Case 1 is a conductive dike under a less conductive near-surface layer. The main targets for Case 2 are two structures in close proximity. To generate the synthetic observations, we use the same survey configuration and mesh design with 5$\%$ Gaussian noise. The results with different noise levels can be found in the supplementary materials. By comparing the inversion result from the conventional and DIP-Inv methods, we will show that the DIP-Inv method provides a useful smoothness effect to the predictions of the boundaries of anomalies (Case 1) and can better distinguish the compact anomalies (Case 2). Due to the fundamental difference in the underlying physics, DC resistivity inversion results typically have lower resolution than FWI results and need more careful design of regularization. $l_2$ norms always promote more blurred structures compared to sparse norms \cite{ref36}; therefore, we choose the results from the conventional method with sparse norms as benchmarks. 

\subsection{Survey Configuration and Mesh Design}
\label{sec:3.1}
We choose a dipole-dipole survey geometry with a station separation of 25 meters and a maximum of 24 receivers per transmitter. The survey line is 700 meters long, so the total number of potential difference measurements is 348. 

\begin{figure}[h!]
\centering
\includegraphics[width=3.5in]{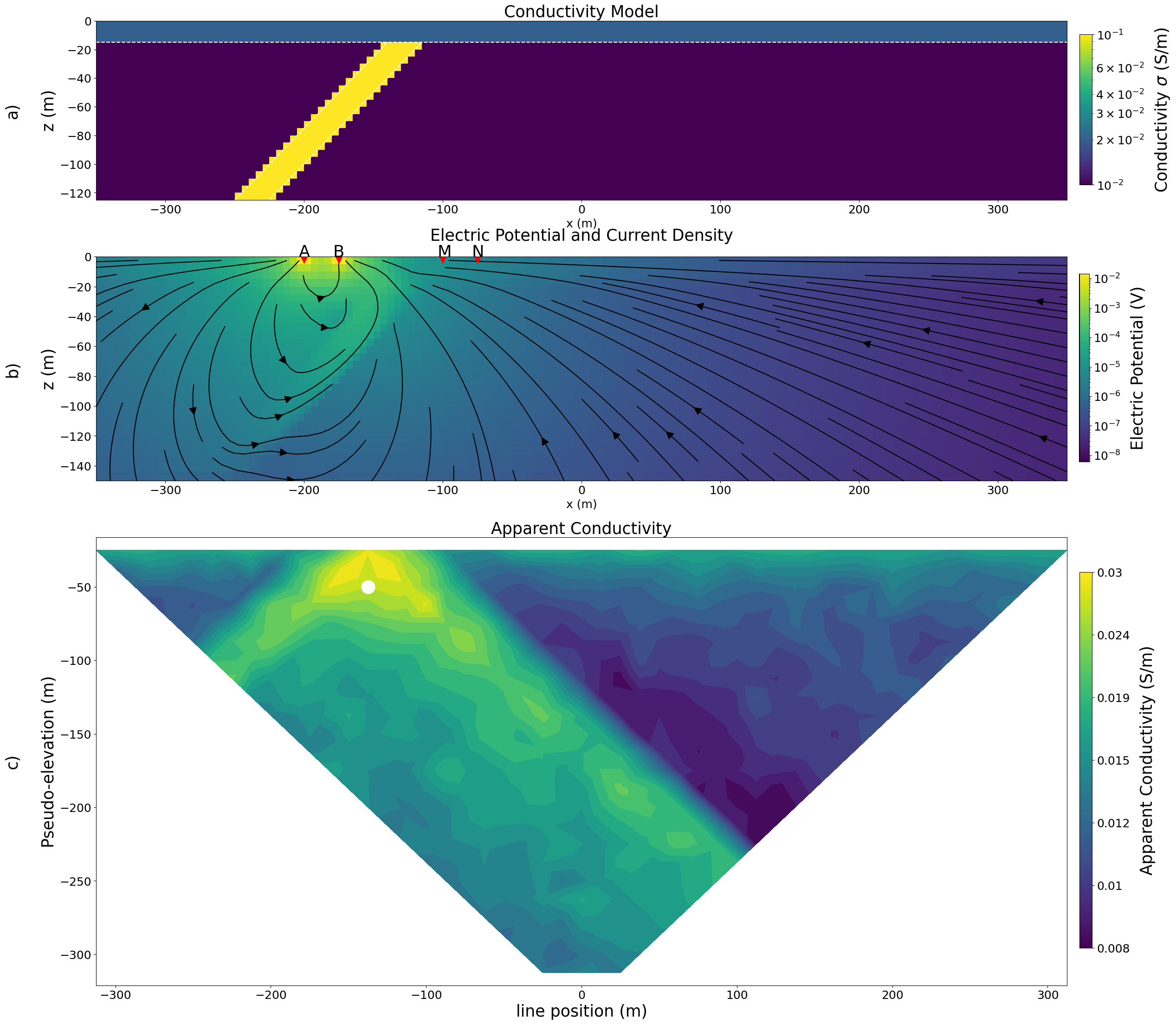}
\caption{(a) is the true conductivity model for Case 1. The top layer has a conductivity of 0.02 S/m. The dike, with a dip angle of 45 degrees, has a conductivity of 0.1 S/m. The background has a conductivity of 0.01 S/m. (b) shows the electrical potential and current density with the 6th transmitter for Case 1. (c) shows the pseudosection for Case 1.}
\label{fig3_1}
\end{figure}

\begin{figure}[h!]
\centering
\includegraphics[width=3.5in]{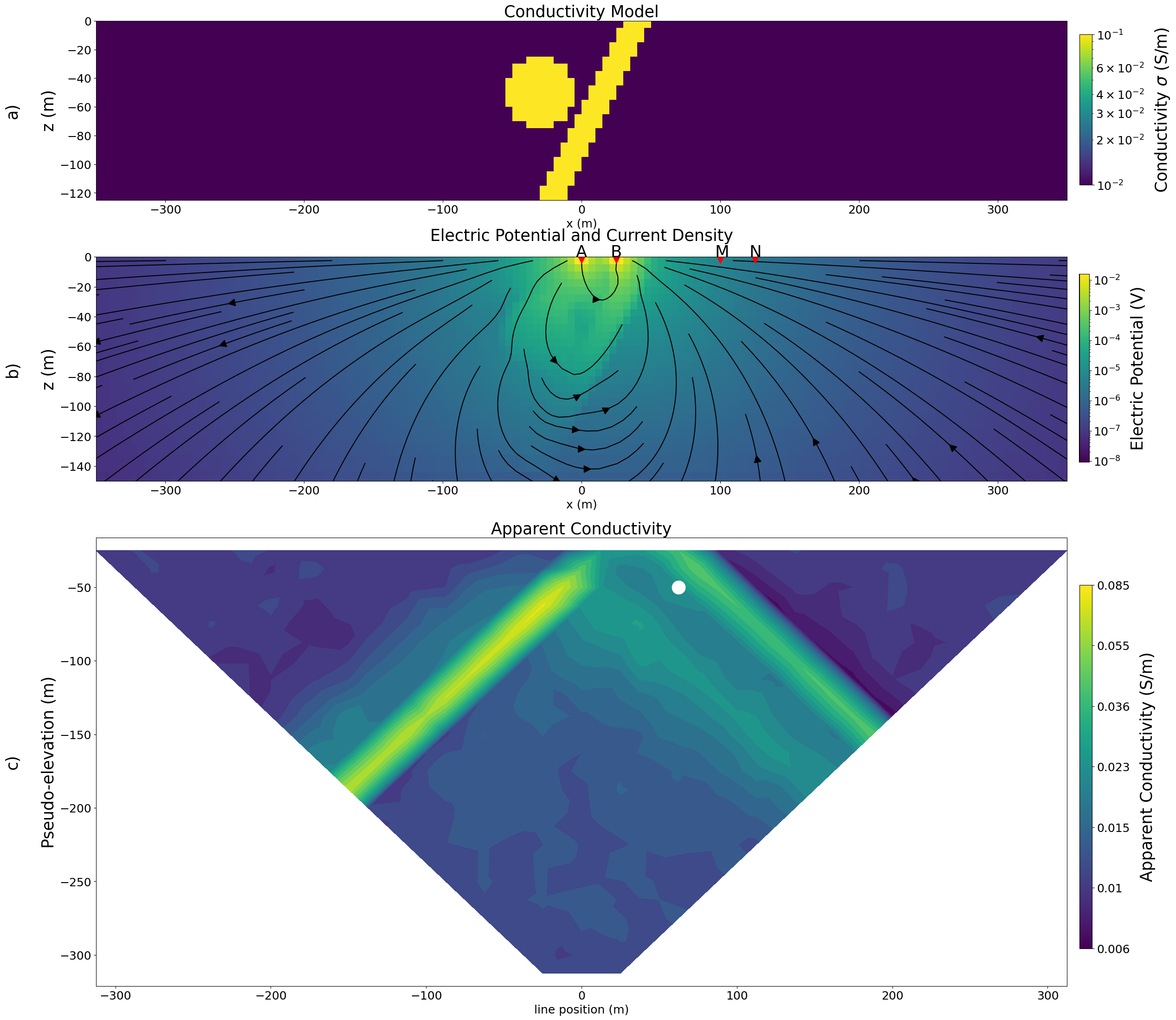}
\caption{(a) is the true conductivity model for Case 2. The cylinder and the dike have a conductivity of 0.1 S/m. The background has a conductivity of 0.01 S/m. (b) shows the electrical potential and current density with the 14th transmitter for Case 2. (c) shows the pseudosection for Case 2.}
\label{fig3_2}
\end{figure}

For the simulation, we use a mesh that has core cells of 5m $\times$ 5m, with 200 core-mesh cells in the x-direction and 25 cells in the z-direction. We add 7 padding cells that extend by a factor of 1.5 to the sides and bottom of the mesh. In total, the mesh has $W\times H = 6944$ cells. The total number of trainable weights in $L$ is 23055 based on the DIP-Inv architecture [Table. \ref{table:1}], so we over-parameterize the problem. We also add a dropout layer to impose a stronger net regularization effect on this over-parameterized problem (see section \ref{sec:4.4} for further discussion). The subsurface conductivity models for Case 1 and Case 2 are shown in Fig. \ref{fig3_1}(a) and Fig. \ref{fig3_2}(a). To show how the current density changes due to the appearance of the anomalies, Fig. \ref{fig3_1}(b) and Fig. \ref{fig3_2}(b) illustrate the currents (streamlines) and resultant electric potentials (background colors) that are obtained by solving equation \ref{Poisson equation} for a single source with the positive electrode at location A and the negative electrode at location B. The positive electrode and negative electrode of the transmitter are labelled as A and B, respectively. The two electrodes of a receiver are labelled as M and N, respectively. We only demonstrate the location of one transmitter-receiver pair, the location of the other transmitter-receiver pairs can be referred from the survey configuration described above. The changes in current density are recorded in the voltage measurements on the ground. The voltage measurements are profiled on the pseudosections [Fig. \ref{fig3_1}(c) and Fig. \ref{fig3_2}(c)]. The pseudosection plots the apparent conductivity, calculated by a voltage measurement and the geometric factor which depends on the survey geometry, at the midpoint of the corresponding transmitter-receiver pair and at the depth that depends on the separation of the transmitter and receiver \cite{ref36}. By measuring the potential difference between electrodes M and N shown in Fig. \ref{fig3_1}(b) or Fig. \ref{fig3_2}(b), we obtain a single datum that is shown as the white dot in the pseudosection. Although the pseudosection does not show a true geological crosssection, it is a useful way to visualize the measurements. 

\subsection{Choice of Optimization Method}
\label{sec:3.2}
In the DIP-Inv method, we choose Adam, a first-order optimizer, with a learning rate of 0.0001 to update weights $w$ in $L$. Note that we only have one input "sample", which is a non-trainable 1 by 8 tensor, so the objective function is not stochastic in this study. The conventional method uses inexact Gauss Newton \cite{ref28}, a second-order optimization method, to update $m$ since the second-order optimization methods are known to speed up convergence for a variety of inverse problems \cite{ref5}. However, researchers found practical challenges in employing second-order optimization methods to train a NN partly because estimating the Hessian is time-consuming when the number of trainable parameters is large \cite{ref10}. As a result, we employ inexact Gauss Newton to update $m$, whose dimension is 6944, in the conventional method and Adam to update $w$, whose dimension is 23055, in the DIP-Inv method. 
 
\subsection{Case 1}
\label{sec:3.3}
We first find the best conventional inversion results with respect to the true model by a grid search over the hyper-parameter space. Namely, we conduct multiple conventional inversions using different values of  $p_{s,x,z}$, $\beta_s$, $\alpha_{s,x,z}$ with or without sensitivity weighting. 

\begin{figure}[h!]
\centering
\includegraphics[width=3.5in]{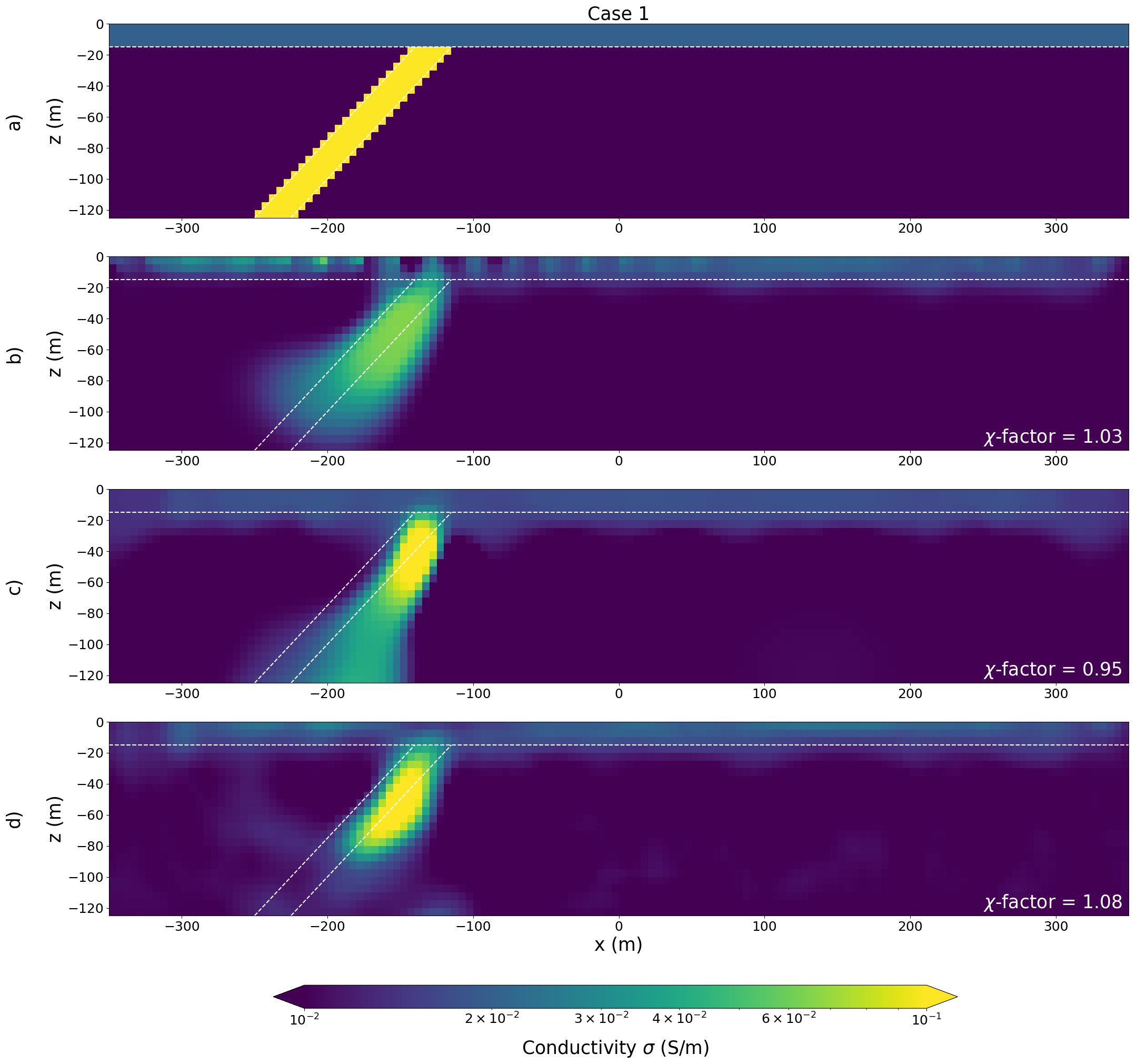}
\caption{(a) is the true conductivity model. (b) and (c) are the conventional inversion results without/with sensitivity weighting, respectively. (d) is the DIP-Inv result.}
\label{fig3_3}
\end{figure}

After that, we find that the best conventional inversion results with respect to the true model are from $\beta_s = 1e2$, $p_s = 0$, $p_x = p_z = 1$, $\alpha_s = 0.005$, and $\alpha_x = \alpha_z = 0.5$ [Fig. \ref{fig3_3}(b) and (c)]. We perform the DIP-Inv method with a decay rate of $\tau = 1000$, and a dropout rate of 0.1 after the fourth hidden block [Fig. \ref{fig3_3}(d)].

All models fit the data to a comparable level, and the $\chi$-factor of each model is shown in the bottom right-hand side of each inversion result. Though each shows a conductive, dipping structure, the details differ. By not including sensitivity weighting, we recover a more diffuse structure [Fig. \ref{fig3_3}(b)]. We also see near-electrodes artifacts. Using sensitivity weighting, as shown in Fig. \ref{fig3_3}(c), reduces the near-electrodes artifacts, but the dip is more vertical than the true model. The DIP-Inv result [Fig. \ref{fig3_3}(d)] does not require any sensitivity weighting and shows an improved recovery of the dip of the dike without any near-electrodes artifacts. In a nutshell, the predicted model in Fig. \ref{fig3_3}(b) has a better indication of dip but has near-surface artifacts, whereas Fig. \ref{fig3_3}(c) removes those artifacts but has the wrong dip angle; therefore, each has useful information, but neither of them satisfactory on its own. In contrast, the predicted model in Fig. 6(d) has good dip angle prediction and no near-electrodes artifacts.

\begin{figure*}
\begin{center}
    \centering
    \includegraphics[width = 1.0\linewidth]{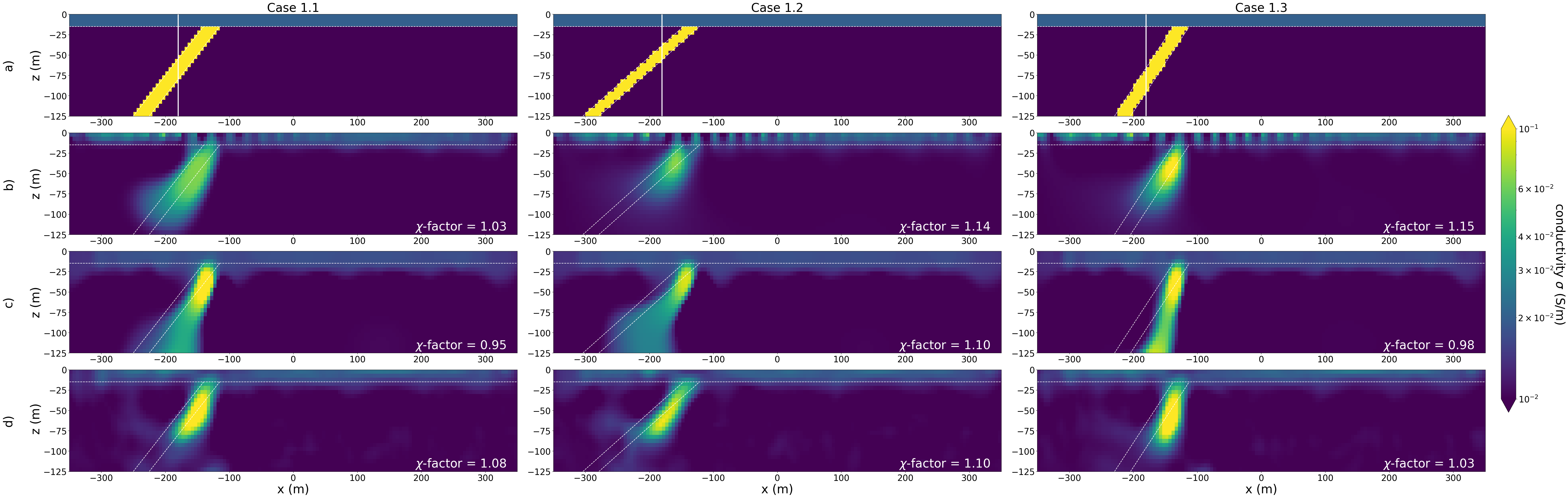}
    \caption{(a) is the true model; the dip of the dike is varied. (b) and (c) are from the conventional sparse-norms inversions without/with sensitivity weighting respectively. (d) is the DIP-Inv result.}
    \label{fig3_4}
\end{center}
\end{figure*}

\begin{figure}[h!]
\centering
\includegraphics[width=3.5in]{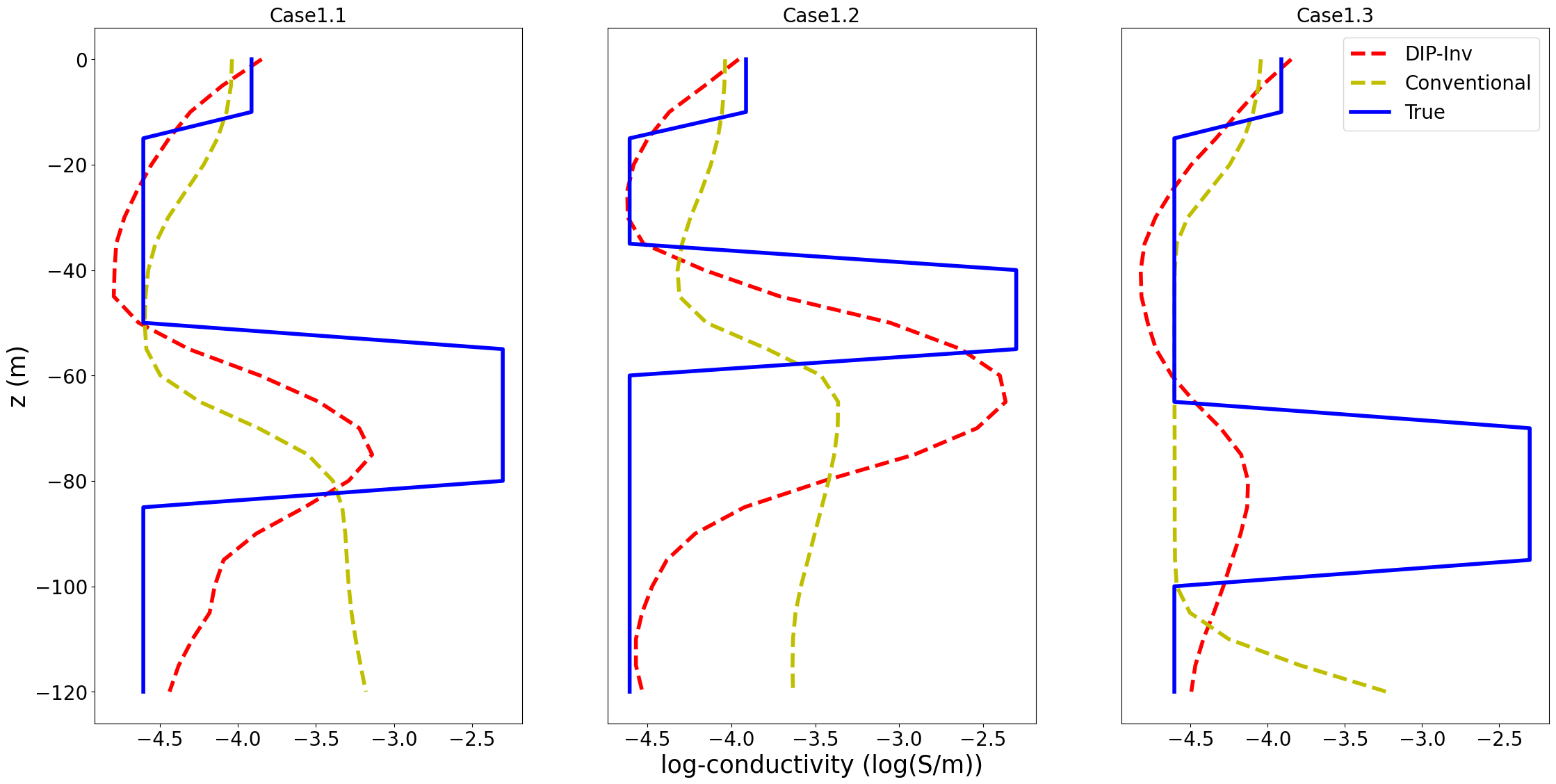}
\caption{The log conductivity values along the vertical profile at x = -180m for Case 1.1-1.3. The locations of these profiles are marked by the white vertical lines in Fig. \ref{fig3_4}(a). Only conventional results from the inversions with sensitivity weighting are shown in the plot.}
\label{fig3_4.5}
\end{figure}

To further test the recovery of the dip angle, we perform two other sets of inversions with two different true models that have different dip angles (See Case 1.2 and Case 1.3 in Fig. \ref{fig3_4}). In the conventional inversion results without using sensitivity weighting, anomaly predictions on the top layer concentrate near the electrodes, and the conductivity values predicted in the dike locations are lower than the true values. The conventional inversion results using sensitivity weighting predict unwanted diffuse structures on the bottom of the dike, especially for Case 1.2. In comparison, the DIP-Inv method recovers the top layer well and surpasses the conventional method in terms of recovery of the dip angle. The conductivity values along a depth profile in Fig. \ref{fig3_4.5} show that the inversion result from the DIP-Inv method follows a more similar trend as the true model compared to the inversion result from the conventional method in all 3 cases: The DIP-Inv method improves the recovery of the position of the dike as compared to the conventional method where it is recovered in a deeper position than the true model. We also show the mean absolute error (MAE) and the mean squared error (MSE) between the inversion results and the true models [Table. \ref{table:7}], which are evaluation metrics that are common in the ML community to compare the performance of algorithms. For the conventional method, we show the scores for the models with sensitivity weighting. The DIP-Inv method has better performance in both metrics on all three models (Case 1.1, 1.2, and 1.3). Considering that the inversion results from the conventional method and the DIP-Inv method have similar chi-factor values, we conclude that the DIP-Inv method produces better recovery of the subsurface structures than the conventional method when both inversions fit observations at the same level. 

\begin{table}[b!]
\centering
\caption{Comparison of the DIP-Inv method and the conventional method with sensitivity weighting}
\begin{tabular}{|c|c|l|l|}
\hline
Model & Metrics & Conventional method & DIP-Inv\\
\hline
Case 1.1 & MAE $\downarrow$ & 0.1862 & \textbf{0.1500}\\
 & MSE $\downarrow$ & 0.0058 & \textbf{0.0051}\\
\hline
Case 1.2 & MAE $\downarrow$ & 0.1832 & \textbf{0.1524}\\
 & MSE $\downarrow$ & 0.0057 & \textbf{0.0051}\\
\hline
Case 1.3 & MAE $\downarrow$ & 0.1691 & \textbf{0.1564}\\
 & MSE $\downarrow$ & 0.0059 & \textbf{0.0054}\\ 
\hline
\end{tabular}
\label{table:7}
\end{table}

\subsection{Case 2}
\label{sec:3.4}
\begin{figure}[h!]
\centering
\includegraphics[width=3.5in]{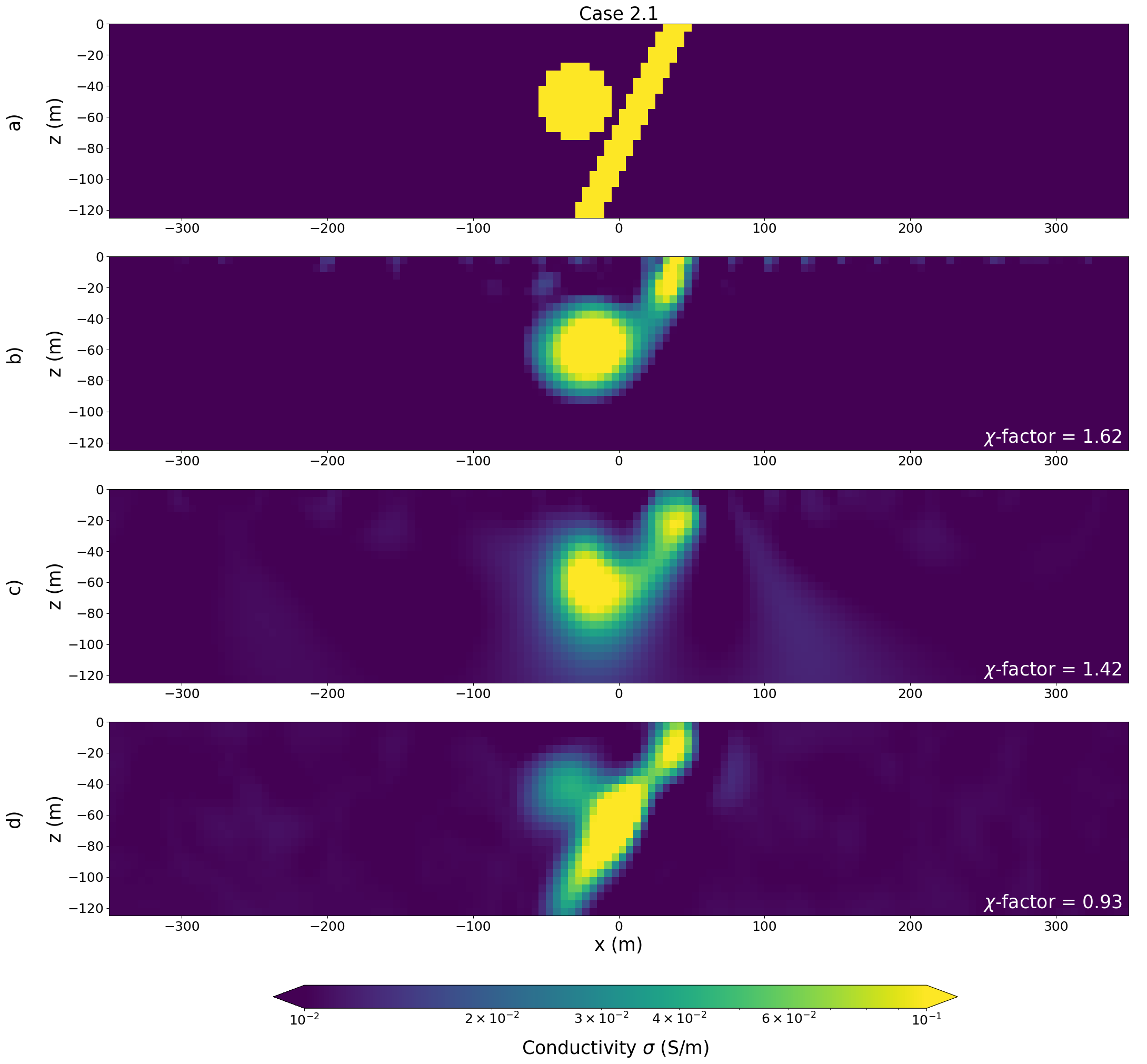}
\caption{(a) is the true model. (b) and (c) are from the conventional sparse-norms inversions without/with sensitivity weighting respectively. (d) is the DIP-Inv result.}
\label{fig3_5}
\end{figure}

After a grid search over the hyper-parameter space, we find that $\beta_s = 1e1$, $p_s = 0$, $p_x = p_z = 2$, $\alpha_s = 0.005$, and $\alpha_x = \alpha_z = 0.5$ give the best conventional results with respect to the true model in Case 2 [Fig. \ref{fig3_5}(b), (c)]. We then perform the DIP-Inv method with decay rate $\tau = 800$, and a dropout rate of 0.05 after the fourth hidden block [Fig. \ref{fig3_5}(d)]. 

Again, all models fit the data to a comparable level. In Fig. \ref{fig3_5}(b) and (c), the predicted models recover the cylindrical structure and a thin dike-like structure near the surface, but the structure of the dike doesn't extend to depth. In contrast, the DIP-Inv result [Fig. \ref{fig3_5}(d)] recovers the dike, and it is clear that it extends at depth, we also see the cylindrical target being recovered at the correct location, though its amplitude is reduced as compared to the true model. 

\begin{figure}[h!]
\centering
\includegraphics[width=3.5in]{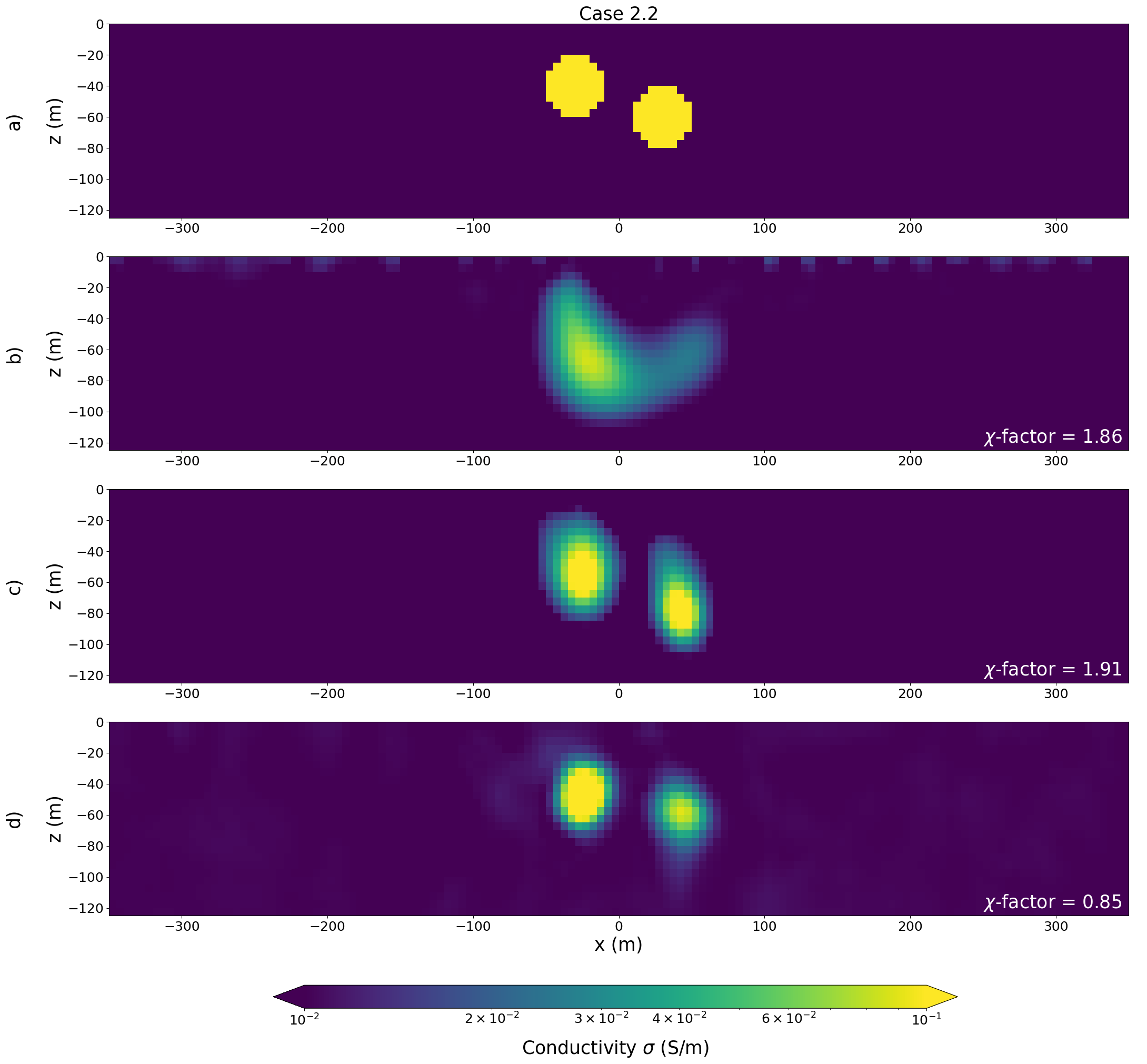}
\caption{(a) is the true model. (b) and (c) are from the conventional sparse-norms inversions without/with sensitivity weighting respectively. (d) is the DIP-Inv result.}
\label{fig_7.5}
\end{figure}

To further verify the argument that the DIP-Inv method can distinguish the two compact anomalies in close proximity. We conduct another trial on a model with two compact, cylindrical targets [Fig. \ref{fig_7.5}a]. 
Using the conventional approach without sensitivity weighting [Fig. \ref{fig_7.5}(b)], we recover a blurred model that connects the two structures. Using sensitivity weighting is an improvement and we can see that there are two distinct bodies. They are somewhat smeared vertically [Fig. \ref{fig_7.5}(c)]. The DIP-Inv result arguably provides the best recovery of the two compact targets [Fig. \ref{fig_7.5}(d)]. 

The MAE and MSE for Cases 2.1 and 2.2 are shown in the Table \ref{table:A}. For Case 2.1 the DIP-Inv method has a better MAE score than the conventional method, and the MSE results are comparable. For Case 2.2, the difference of both scores are within 20$\%$. However, when visually inspecting these models, the DIP-Inv results would arguably lead to a geological interpretation that is closest to the true model. For Case 2.1, we can clearly see an extended, dipping conductor. In Case 2.2, both the conventional method with sensitivity weighting (Fig. \ref{fig_7.5}(c)) and the DIP-Inv result (Fig. \ref{fig_7.5}(d)) resolve the two isolated, cylindrical targets. A benefit of the DIP-Inv method is that no sensitivity weighting was necessary, which shows the utility of the implicit regularization from the CNN. 

\begin{table}[h!]
\centering
\caption{Comparison of the DIP-Inv method and the conventional method with sensitivity weighting}
\begin{tabular}{|c|c|l|l|}
\hline
Model & Metrics & Conventional method& DIP-Inv\\
\hline
Case 2.1 & MAE $\downarrow$ & 0.1313 & \textbf{0.0997}\\
 & MSE $\downarrow$ & \textbf{0.0043} & 0.0047\\
\hline
Case 2.2 & MAE $\downarrow$ & \textbf{0.0533} & 0.0580\\
 & MSE $\downarrow$ & 0.0039 & \textbf{0.0033}\\
\hline
\end{tabular}
\label{table:A}
\end{table}

\section{Implicit Regularization in DIP-Inv}
\label{sec:4}
In this section, we will explore the implicit regularization effect observed in the results from the DIP-Inv method. We hypothesize that the smoothing effect generated by the bi-linear upsampling operator is a key component of the DIP-Inv approach. We then test this hypothesis by replacing bi-linear upsampling with other upsampling operators. Finally, we present the effect of changing the number of hidden blocks and adding dropout layers. 

\subsection{Bi-linear Upsampling Operator}
\label{sec:4.2}
In this subsection, we test the hypothesis that the observed implicit regularization effect is mainly from the bi-linear upsampling operator. Bi-linear upsampling is a common choice for up-scaling the dimension of certain hidden layers. If we consider that the i-th hidden layer is a bi-linear upsampling layer with upsampling scale $k$ and input dimension $(a, b, c, d)$, then the output dimension for the i-th hidden layer is $(a, b, kc, kd)$. The bi-linear interpolation is employed in this up-scaling process. 

\begin{figure}[!t]
\centering
\includegraphics[width=3.5in]{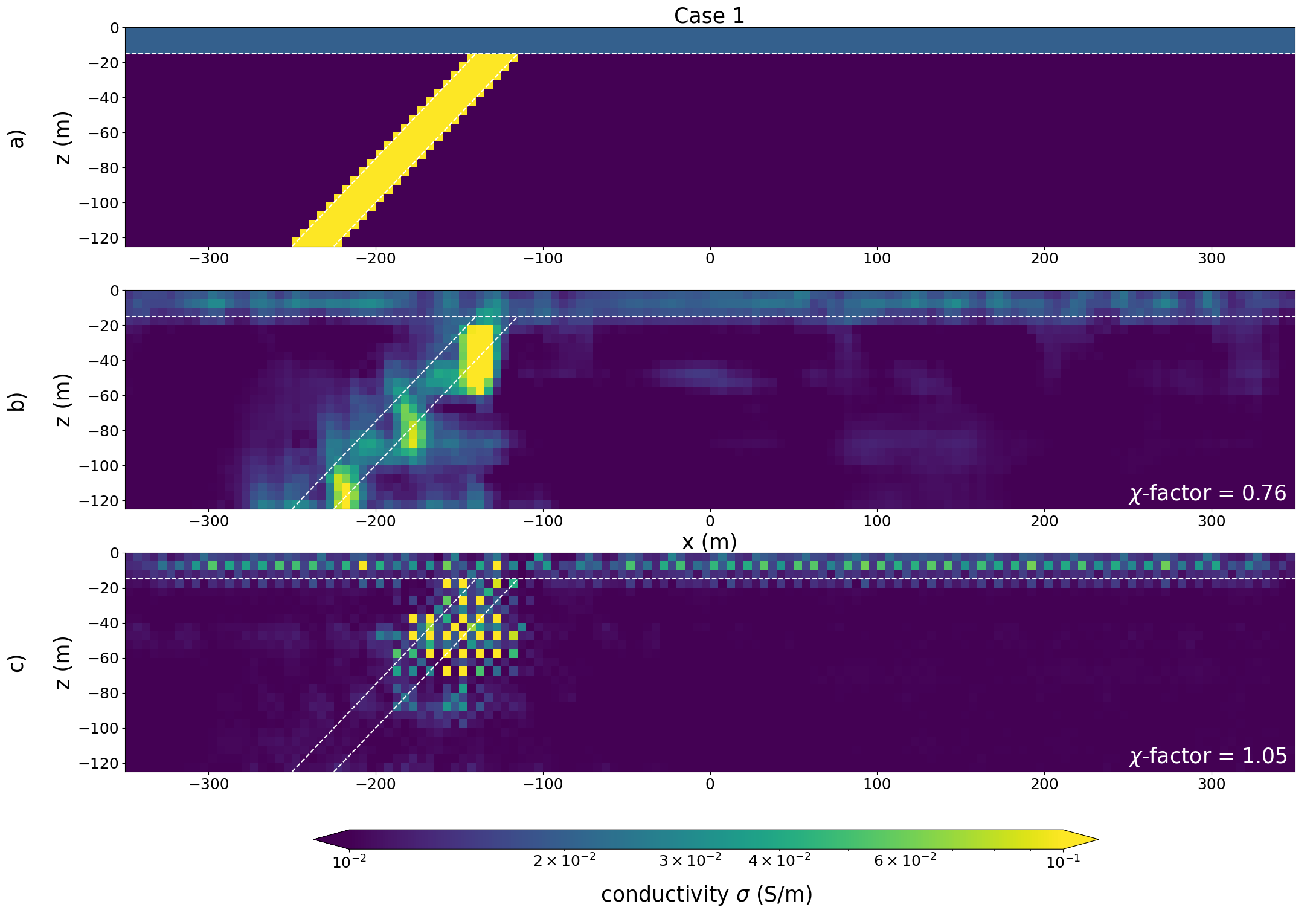}
\caption{(a) is the true model. (b) and (c) are from inversions using nearest neighbour upsampling and transposed convolution operator respectively.}
\label{fig3_9}
\end{figure}

\begin{figure}[!t]
\centering
\includegraphics[width=3.5in]{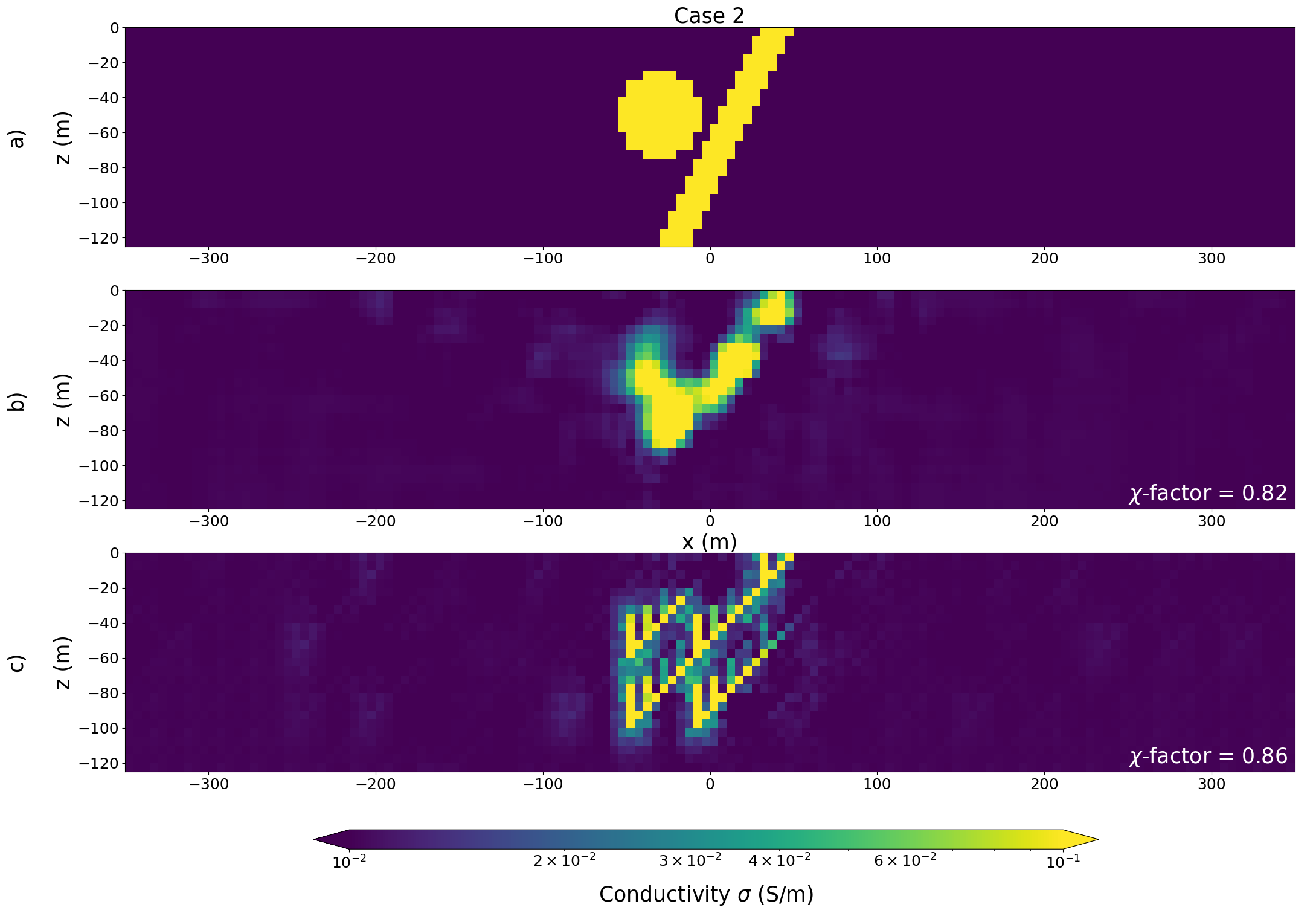}
\caption{(a) is the true model. (b) and (c) are from inversions using nearest neighbour upsampling and transposed convolution operator respectively.}
\label{fig3_10}
\end{figure}

The output value of the bi-linear interpolation in each pixel is a distance-based, weighted sum of the surrounding pixels; therefore, bi-linear interpolation generates a pixel-level smoothed output. To verify our hypothesis, we replace the bi-linear upsampling with the nearest neighbour upsampling or the transposed convolution. Our experiments demonstrated that these alternatives yield inferior results [Fig. \ref{fig3_9} and \ref{fig3_10}], which supports that the bi-linear upsampling is critical in the success of DIP-Inv.

\subsection{Influence of Number of Layers}
\label{sec:4.3}
To further test our hypothesis that the implicit regularization effect is partly from the bi-linear upsampling operator, we show the results from changing the number of bi-linear upsampling layers by two. If our hypothesis is true, then the regularization effect should be amplified as we increase the number of hidden blocks (one hidden block has one bi-linear upsampling layer), and vice versa. Similar to the need to choose a good trade-off between the data misfit term and the regularization term in the conventional method, we need to choose how much regularization effect we want to impose for our DIP-Inv method on a case-by-case basis. Too much or too little regularization is both problematic. 

\begin{figure}[!t]
\centering
\includegraphics[width=3.5in]{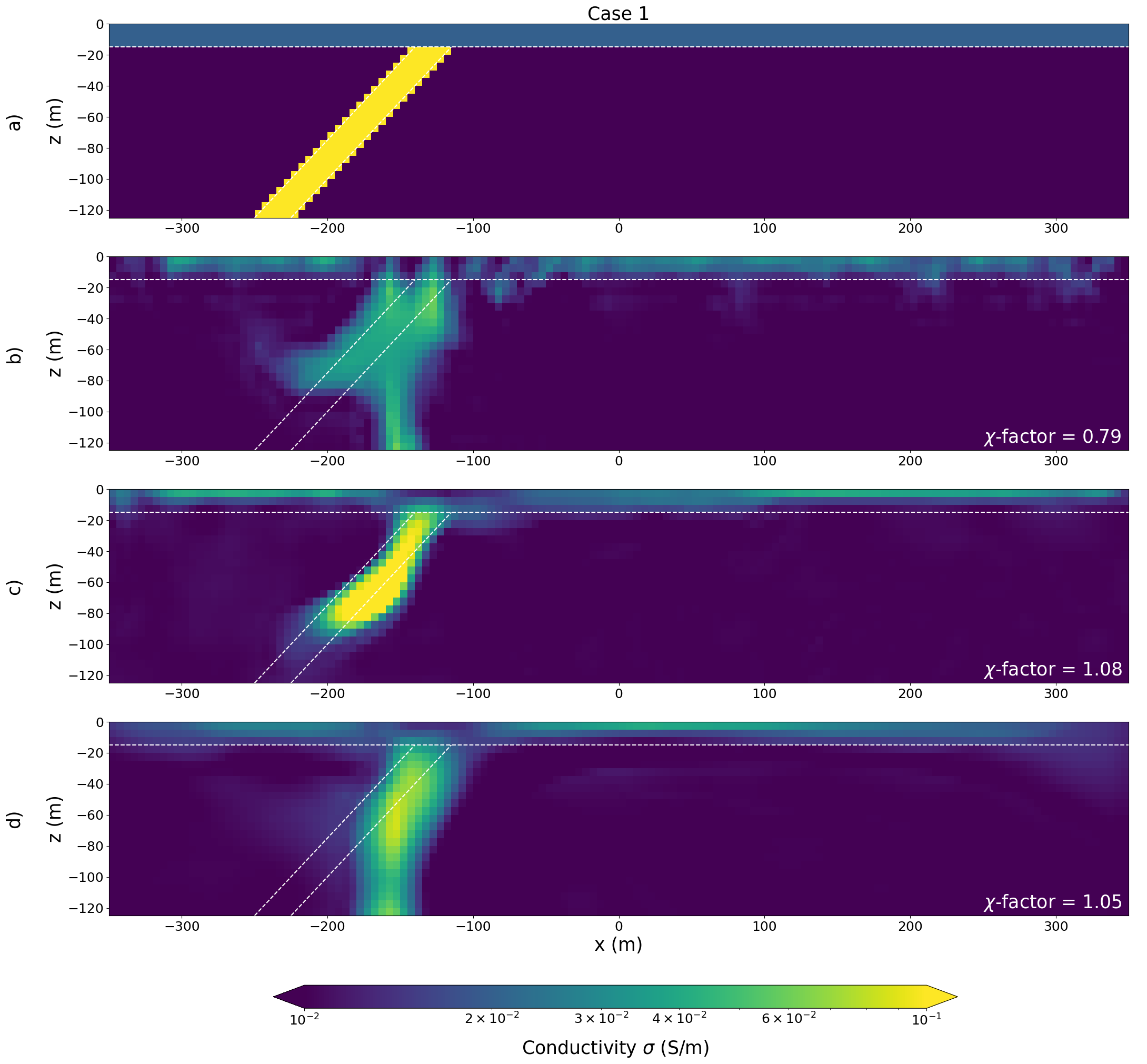}
\caption{(a) is the true model. (b), (c), and (d) are results for employing 1, 3, and 5 hidden blocks respectively. All results are from the CNN architectures without any dropout layer.}
\label{fig3_12}
\end{figure}

\begin{figure}[!t]
\centering
\includegraphics[width=3.5in]{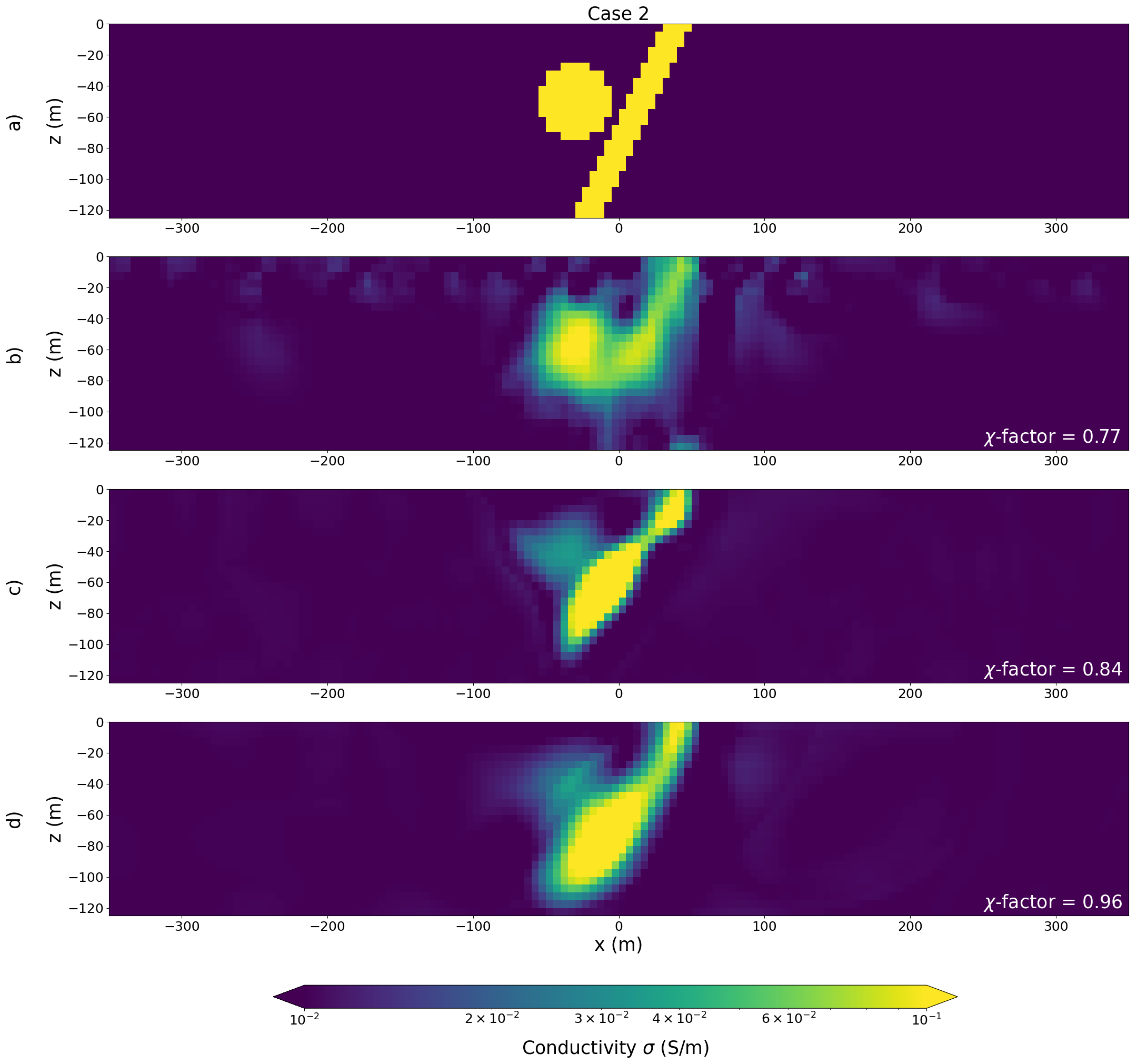}
\caption{(a) is the true model. (b), (c), and (d) are results for employing 1, 3, and 5 hidden blocks respectively. All results are from the CNN architectures without any dropout layer. }
\label{fig3_13}
\end{figure}

The DIP-Inv inversion results for Case 1 using 1, 3 and 5 hidden blocks are shown in Fig. \ref{fig3_12}. 
We don't use a dropout layer for all experiments presented in this subsection to avoid ambiguity on the source of the regularization effect. 
Using only one hidden block [Fig. \ref{fig3_12}(b)], we recover a model with lots of structure and artifacts, illustrating that one hidden block is insufficient for providing enough regularization effect.
On the other hand, when we use five hidden blocks, we recover a more blurred structure [Fig. \ref{fig3_12}(d)] because the additional hidden blocks provide too much smooth effect. 
For this model, we find that using three hidden blocks [Fig.\ref{fig3_12}(c)] provides an appropriate level of regularization. 
Note that the optimal number of layers will depend on the subsurface model. 
We repeat this experiment for the model from Case 2 to support our conclusion and show these results in Fig. \ref{fig3_13}. Again, we see that increasing the number of hidden blocks increases the regularization effect and using three layers provides an appropriate level of regularization. 

\subsection{Enhanced Smoothing Effects from the Dropout Layer}
\label{sec:4.4}
\begin{figure}[h!]
\centering
\includegraphics[width=3.5in]{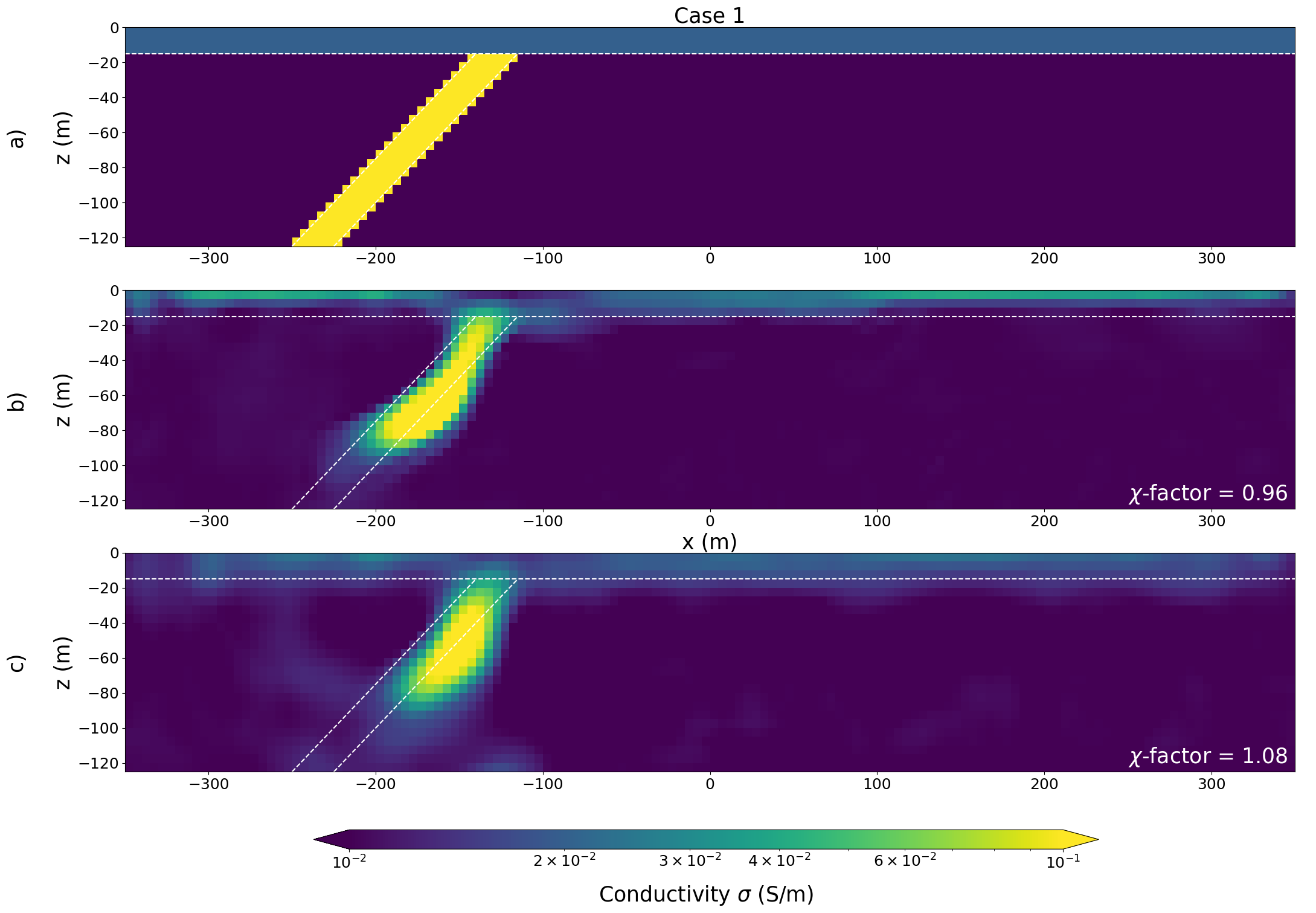}
\caption{(a) is the true model. (b) is the inversion result from the DIP-Inv methods without any dropout layer, and (c) is the inversion result from the DIP-Inv methods with a dropout layer.}
\label{fig3_14}
\end{figure}

\begin{figure}[h!]
\centering
\includegraphics[width=3.5in]{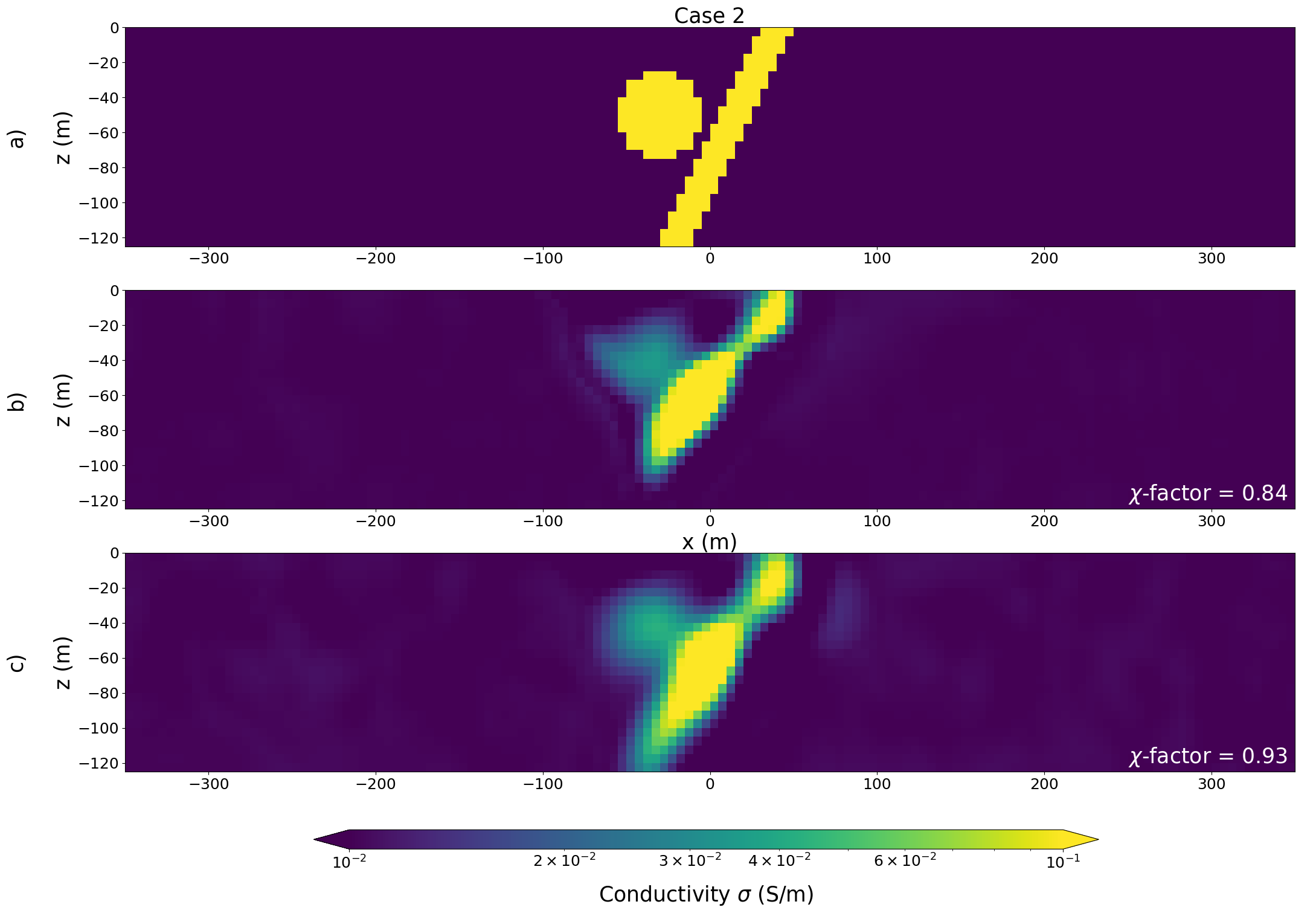}
\caption{(a) is the true model. (b) is the inversion result from the DIP-Inv methods without any dropout layer, and (c) is the inversion result from the DIP-Inv methods with a dropout layer.}
\label{fig3_15}
\end{figure}

Dropout is used to prevent over-fitting in training large neural networks \cite{ref23}. It surpasses conventional regularization methods such as Tikhonov regularization and max-norm regularization in supervised tasks in computer vision, speech recognition, etc \cite{ref23}. When using dropout, we randomly zero out certain neurons during each training iteration. Since different neurons are chosen to be zero-outed in each iteration, this is a stochastic regularization method. Intuitively, this stochastic zero-out scheme prevents the final output from relying too much on any particular neuron. Therefore, it improves the generalization of the neural network. Dropout can be viewed as a model sampling method as well. In each iteration, it will sample a new CNN model with different zero-outed neurons compared to the model in the last iteration, so the output of the last iteration would be approximately a weighted average of different models \cite{ref23}. See Fig. \ref{fig3_14} and \ref{fig3_15} for an ablation study on having/not having a dropout layer in Case 1 and 2, respectively. We choose a dropout rate of 0.1 and only add 1 dropout layer after the fourth hidden block of the DIP-Inv. 

The dropout layer is useful for improving the recovery of the top layer of the subsurface for Case 1 as shown in Fig. \ref{fig3_14} and recovery of the cylindrical target for Case 2 as shown in Fig. \ref{fig3_15}. A practical guide for applying the DIP-Inv would be to perform it without any dropout layer in the first attempt, and then perform it with the dropout layer if the first result is not satisfying.  

\section{Future Work}
The above analyses are mainly based on experimental results, a more rigorous mathematical explanation is a natural direction of the future work. The mathematical explanation for the implicit bias of neural networks is still an open question. For example, Frei et al. \cite{ref48} proved that a 2-layer Leaky ReLU MLP is a $l_2$-max-margin solution under certain conditions. Rahaman et al. \cite{ref49} found that ReLU NN is biased toward learning lower-frequency variation. Some scholars have studied the implicit bias from the gradient-based optimization methods to answer the generalization puzzle in NN \cite{ref50}. Specifically, for convolutional neural networks, Tachella et al. \cite{ref61} stated that Adam is key to the success of utilizing the DIP method in image denoising. These explanations are limited by the assumptions they made to prove the relevant theorems and do not perfectly interpret the implicit regularization effects that we have observed in DIP-Inv. However, these works shed light on explaining the good performance of over-parameterized neural networks. 

The idea of reparameterizing the mesh space can be adapted to other Tikhonov-style geophysical inversion problems (e.g. potential fields and electromagnetics, where SimPEG has other simulations that could be swapped in). This method can also be combined with additional explicit regularization terms such as using a non-uniform reference model based on the petrophysical knowledge for the survey domain. Currently, the DIP-Inv method requires more than 1000 iterations, and it is much slower than a conventional inversion, so exploring more efficient optimization methods is also a direction of future research. Moreover, to illustrate the impacts of implicit regularization on models with known structures, we only show the synthetic cases. Future work could also include examining the results of field data. Similarly, we chose to work with a simple CNN architecture, but exploring more complex architectures would be an avenue for further research.

\section{Conclusion}
Motivated by the works of utilizing the implicit regularization inherently included in the CNN structure in the inverse problems in other fields, we explore the utility of this implicit regularization in the Tikhonov-style geophysical inversions. In this study, we focus on the DC resistivity inversion with synthetic examples and compare the DIP-Inv results with the conventional Tikhonov-regularized inversion results using the sparse norms. Compared to the conventional method, the DIP-Inv method improves the recovery of the dip of a target and the recovery of the compact targets in close proximity. We also show that the bi-linear upsampling operator is a key component in DIP-Inv and adding dropout can be beneficial in some cases. Those results illustrate that the implicit regularization from the CNN structure can be useful for the Tikhonov-style geophysical inversion. As a test-time learning method, DIP-Inv can be adapted to other geophysical inversion problems or combined with explicit regularization terms without collecting a training dataset.

\section*{Data Availability}
The code used is available at \\https://doi.org/10.5281/zenodo.10289234.

\section*{Acknowledgments}
This research was supported by funding from the NSERC Discovery Grants Program and the Mitacs Accelerate Program. We thank Jorge Lopez-Alvis for his comments and suggestions on paper writing. We would also like to thank Doug Oldenburg, Devin Cowan, Joseph Capriotti, John M. Weis, Johnathan C. Kuttai, Jingrong Lin and Santiago Soler for their invaluable guidance and advice on the usage of SimPEG.

\vspace{-33pt}

\begin{IEEEbiography}[{\includegraphics[width=1in,height=1.25in,clip,keepaspectratio]{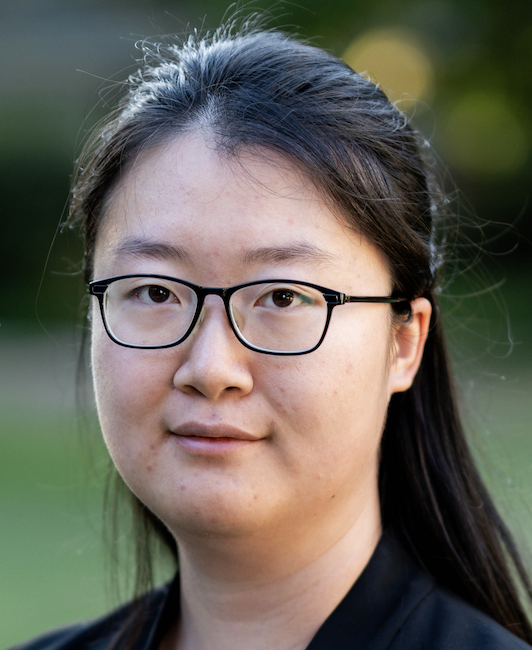}}]{Anran Xu}
received the Honours Bachelor of Science degree in Mathematics $\&$ Its Applications Specialist (Physical Science) and Physics Major from the University of Toronto, Toronto, ON Canada, in 2022. She is currently pursuing a Master of Science in Geophysics with the University of British Columbia, Vancouver, BC Canada. 
\par
Her research interests include inverse problems and machine learning applications. 
\end{IEEEbiography}
\begin{IEEEbiography}
[{\includegraphics[width=1in,height=1.25in,clip,keepaspectratio]{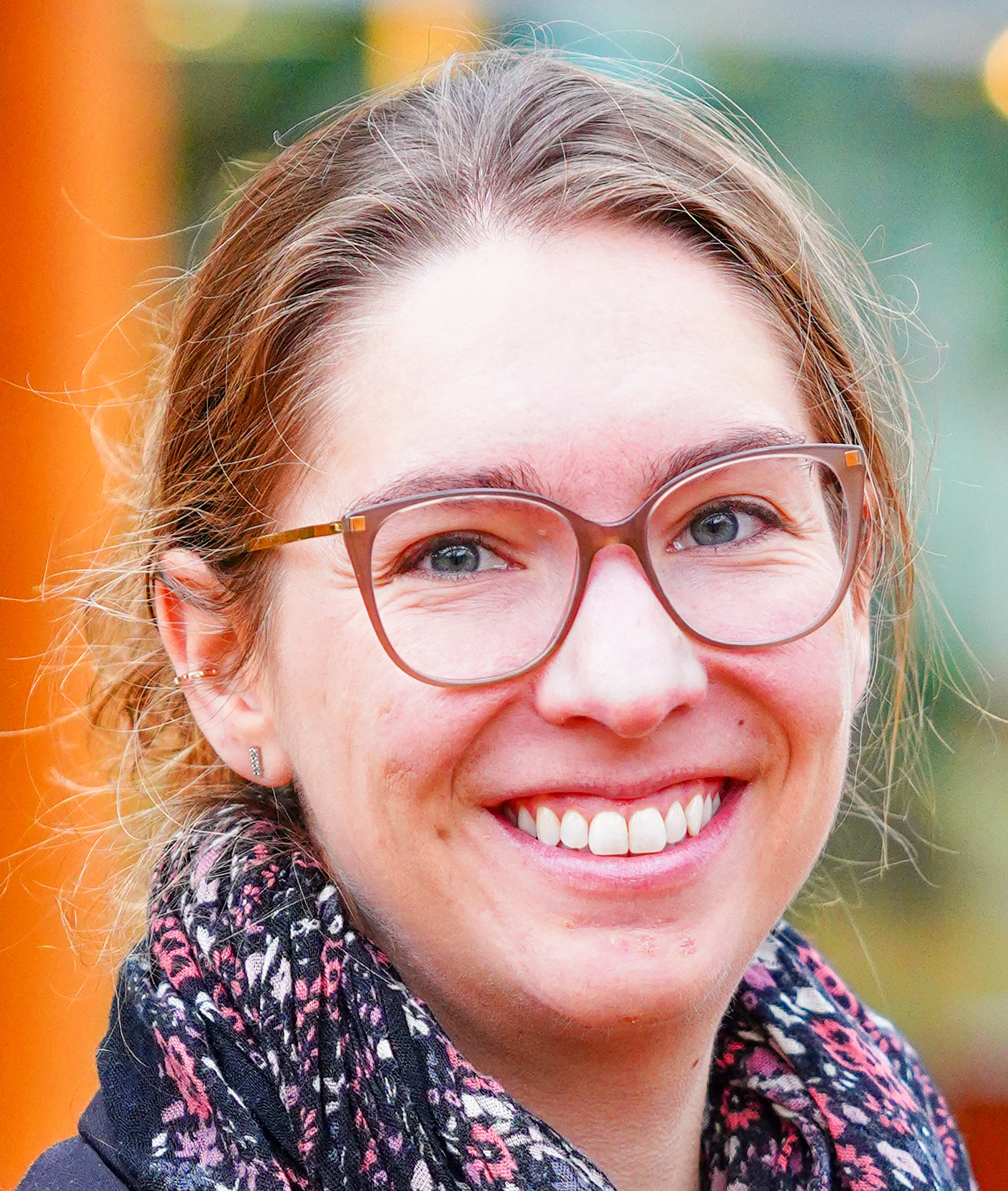}}]{Lindsey J. Heagy} is an Assistant Professor in the Department of Earth, Ocean and Atmospheric Sciences and Director of the Geophysical Inversion Facility at UBC. She completed her BSc in geophysics at the University of Alberta in 2012 and her PhD at UBC in 2018. Prior to her current position, she was a Postdoctoral researcher in the Statistics Department at UC Berkeley. 
\par
Her research combines computational methods in numerical simulations, inversions, and machine learning to characterize the geophysical subsurface.

\end{IEEEbiography}

\vfill

\end{document}